\documentclass[sigconf]{acmart}
\AtBeginDocument{%
  \providecommand\BibTeX{{%
    \normalfont B\kern-0.5em{\scshape i\kern-0.25em b}\kern-0.8em\TeX}}}

\usepackage{adjustbox}
\usepackage{colortbl}
\usepackage{pgfplots}
\usepackage{multirow}
\definecolor{Gray}{gray}{0.88}
\definecolor{DarkGray}{gray}{0.8}
\definecolor{LightGray}{gray}{0.93}
\definecolor{Green}{RGB}{95,150,105}
\definecolor{Purple}{RGB}{122,95,156}
\definecolor{LightBlue}{RGB}{99,150,216}
\definecolor{LightRed}{RGB}{217,133,148}
\pgfplotsset{compat=1.8}
\usepgfplotslibrary{statistics}
\usepackage{color,soul}

\usepackage{dblfloatfix} 

\begin{document}

\acmYear{2023}\copyrightyear{2023}
\setcopyright{acmlicensed}
\acmConference[ICCPS '23]{ACM/IEEE 14th International Conference on Cyber-Physical Systems (with CPS-IoT Week 2022)}{May 9--12, 2023}{San Antonio, TX, USA}
\acmBooktitle{ACM/IEEE 14th International Conference on Cyber-Physical Systems (with CPS-IoT Week 2022) (ICCPS '23), May 9--12, 2023, San Antonio, TX, USA}
\acmPrice{15.00}
\acmDOI{10.1145/3576841.3585933}
\acmISBN{979-8-4007-0036-1/23/05}

\title{Pishgu: Universal Path Prediction Network Architecture for Real-time Cyber-physical Edge Systems}

\author{Ghazal Alinezhad Noghre}
\email{galinezh@uncc.edu}
\affiliation{%
  \institution{University of North Carolina at Charlotte}
  \state{North Carolina}
  \country{USA}
}
\authornote{Both authors contributed equally to this research.}

\author{Vinit Katariya}
\authornotemark[1]
\email{vkatariya@uncc.edu}
\affiliation{%
  \institution{University of North Carolina at Charlotte}
  \state{North Carolina}
  \country{USA}
}

\author{Armin Danesh Pazho}
\affiliation{%
  \institution{University of North Carolina at Charlotte}
  \city{Charlotte}
  \state{North Carolina}
  \country{USA}}
\email{adaneshp@uncc.edu}

\author{Christopher Neff}
\affiliation{%
  \institution{University of North Carolina at Charlotte}
  \city{Charlotte}
  \state{North Carolina}
  \country{USA}}
\email{cneff@uncc.edu}

\author{Hamed Tabkhi}
\affiliation{%
  \institution{University of North Carolina at Charlotte}
  \city{Charlotte}
  \state{North Carolina}
  \country{USA}}
\email{htabkhiv@uncc.edu}


\begin{abstract}
Path prediction is an essential task for many real-world Cyber-Physical Systems (CPS) applications, from autonomous driving and traffic monitoring/management to pedestrian/worker safety. These real-world CPS applications need a robust, lightweight path prediction that can provide a universal network architecture for multiple subjects (e.g., pedestrians and vehicles) from different perspectives. However, most existing algorithms are tailor-made for a unique subject with a specific camera perspective and scenario. This article presents Pishgu, a universal lightweight network architecture, as a robust and holistic solution for path prediction. Pishgu's architecture can adapt to multiple path prediction domains with different subjects (vehicles, pedestrians), perspectives (bird's-eye, high-angle), and scenes (sidewalk, highway). Our proposed architecture captures the inter-dependencies within the subjects in each frame by taking advantage of Graph Isomorphism Networks and the attention module. We separately train and evaluate the efficacy of our architecture on three different CPS domains across multiple perspectives (vehicle bird's-eye view, pedestrian bird's-eye view, and human high-angle view). Pishgu outperforms state-of-the-art solutions in the vehicle bird's-eye view domain by 42\% and 61\% and pedestrian high-angle view domain by 23\% and 22\% in terms of ADE and FDE, respectively. Additionally, we analyze the domain-specific details for various datasets to understand their effect on path prediction and model interpretation. Finally, we report the latency and throughput for all three domains on multiple embedded platforms showcasing the robustness and adaptability of Pishgu for real-world integration into CPS applications.
\end{abstract}







\maketitle
\pagestyle{plain}
\section{Introduction}
Artificial Intelligence (AI) plays an essential role in many emerging Cyber-physical Systems (CPS), such as smart-video surveillance, traffic management, autonomous driving, anomaly detection, workers' safety, and many more. Path prediction is employed in many real-world computer vision applications \cite{wang2019exploring, georgiou2018moving, 9797078, kaur2022applications}, which need real-time analysis of the subjects and proper proactive decision-making working on CPS with limited resources \cite{9757168}. It predicts the paths of the subjects in a scene/frame based on their movement in the past few seconds. A wide range of computer vision applications for CPS in pedestrian safety, transportation safety, intelligent traffic monitoring, and video surveillance can benefit from an accurate, robust, and efficient path-prediction algorithm. The applications vary from a bird's-eye view for drone-related and environmental monitoring applications to a high-angle view for safety, traffic monitoring, and surveillance application for pedestrians and moving vehicles. However, there is a considerable gap between the research in this area and their real-world deployment on CPS; often, path prediction architectures only target a narrow domain and cannot generalize over different subjects and viewpoints. As a result, predicting the path for each domain, such as vehicle bird's-eye view, pedestrian bird's-eye view, or pedestrian high-angle view, requires a distinct architecture to work accurately. Each architecture has its unique requirements and dependencies, increasing the stress on the edge device's hardware and software sides. On the other hand, the research in this area needs to consider real-world limitations since current designs often have huge model sizes with millions of parameters, making their deployment on edge devices almost impossible. 

Path prediction as a whole has a set of universal objectives. All path prediction algorithms are tasked to determine future trajectories based on the current/past trajectories across all subjects of interest (e.g., vehicles or pedestrians). The algorithms must capture the individual behaviors of all subjects and their complex interactions with respect to each other (social interaction) and the surrounding environment (environmental interaction). Each context/domain often imposes unique requirements and nuances within these similar objectives. For instance, a busy highway has hundreds of vehicles in a scene, and their speeds will be around 30 meters per second \cite{NGSIM_i80, NGSIM_US101}. The camera angles cover a large field of view from a very high angle to capture the fast-moving vehicles early enough to generate accurate predictions. In contrast, a surveillance camera overlooking a campus sidewalk will rarely have over a hundred people who move much slower, around 1.4 meters per second \cite{awad2018benchmarking}. The camera will also be much closer to the ground, making the view less bird's-eye and more high-angle.

We observe a proliferation of algorithms \cite{Pip2021, GSTCN2022, STALSTM2021, yue2022human, mendieta2021carpe} that try to optimize the path prediction for a single context and perspective. Most existing approaches focus on an isolated domain and fail to evaluate the generality of their solution over different contexts and domains \cite{chen2022scept, lv2021ssagcn, GSTCN2022}. It is rare to see approaches that are evaluated on both vehicles and pedestrians \cite{salzmann2020trajectron++, chen2022scept}, and only recently have a few pedestrian-focused approaches started using both bird's-eye and high-angle views \cite{gupta2018social, liang2019peeking, liang2020garden, liang2020simaug, li2022graph}. However, a universal architecture that can be trained and accurately predict pedestrians' and vehicles' paths across different views and perspectives is highly desirable when it comes to real-world deployment and applications.

At the same time, path prediction is inherently a real-time task with a demand for one single accurate prediction for all subjects of interest at any point of time \cite{mendieta2021carpe, 2020gripplus}. However, many existing path predictions rely on a spectrum of predictions with significantly large model sizes that heavily occupy the storage and computational power of edge devices. Most works in this context predict multiple future trajectories and choose the best to assess their accuracy and performance \cite{yue2022human, mangalam2021goals, wong2021view}. Predicting a spectrum of possibilities and picking the best one makes the real-world implementation of such models infeasible and fruitless.

This paper introduces Pishgu, a universal path prediction architecture for various applications. Using a single architecture for multiple domains is a step toward solving a significant challenge in developing CPS applications and ensures better utilization of edge devices' limited storage capacity and computational power. Pishgu borrows Graph Isomorphism Network (GIN) \cite{xu2018powerful} formulation for capturing and modeling the relative information between subjects. In addition to GIN, Pishgu goes further by leveraging efficient attention mechanisms to create robustness against noisy data, a large number of parallel path predictions, and environmental alterations and help the predictor to focus on more informative features. With this, Pishgu sets a new State-of-the-Art (SotA) for two crucial path prediction domains, vehicle bird's-eye view and pedestrian high-angle view. We do not evaluate Pishgu for the vehicle high-angle domain due to the lack of high-angle vehicle trajectory datasets. In the vehicle domain, Pishgu achieves up to 2$\times$ improvement in Root Mean Squared Error (RMSE) and 1.7$\times$ and 2.5$\times$ improvements in Average Displacement Error (ADE) and Final Displacement Error (FDE), respectively, when compared to the current SotA approach. In the pedestrian high-angle view domain, Pishgu can improve ADE and FDE by 1.3$\times$. Additional evaluation on real-world embedded platforms demonstrates Pishgu's suitability for real-time CPS applications, with the sample latency across all domains under four milliseconds. Pishgu achieves between 50 and 190 FPS in the pedestrian domain on embedded platforms.

In summary, this paper has the following contributions:

\begin{itemize}    
    \item  We introduce Pishgu, a universal architecture for path prediction with a novel formulation based on the graph isomorphism network and attention mechanism for a wide range of CPS applications that demand robust, accurate, and lightweight path prediction across different subjects and views.
    
    \item We provide an exhaustive domain analysis across the major datasets and domains in the vehicle and pedestrian path prediction to better understand the variations in Pishgu's accuracy with respect to the richness and characteristics of datasets within each domain.
    
    \item We comprehensively evaluate Pishgu across multiple domains, with respect to each domain's characteristics, and demonstrate Pishgu's SotA accuracy in two major domains: (1) Vehicle bird's-eye view, and (2) Pedestrian high-angle view.
    
    \item To verify Pishgu's suitability for real-time CPS applications, we report the latency and throughput on NVIDIA Jetson Xavier NX and NVIDIA Jetson Nano, low-power embedded platforms commonly used in real-world edge devices.
\end{itemize}

In the remainder of this paper, Section \ref{sec:RelatedWork} reviews the crucial works done in real-time path prediction with their innovations and limitations. Section \ref{sec:pishgu} explains the details of the design of Pishgu and how it is aligned with our goal of introducing a universal architecture for path prediction on CPS edge devices. Section \ref{sec:domain} details the challenges and requirements of each domain, and the statistics of datasets used in this manuscript for evaluation. Section \ref{sec:Experiments} presents the results of our extensive experiments on selected datasets in multiple domains. Section \ref{sec:realtime} demonstrates the performance of Pishgu on multiple embedded platforms suitable for CPS edge applications. Finally, Section \ref{sec:conclusion} concludes by highlighting our major contributions and the future direction.

\vspace*{-0.3cm}

\section{Related Work}\label{sec:RelatedWork}

Deep Learning based prediction architectures are rapidly becoming an integral part of modern edge-based CPS \cite{Zhou2021wide_atten, Isern2020Reconfig, deniz2022efficient}. The Transportation CPS (TCPS) uses deep learning for predicting traffic flow density \cite{Jeong2013traffFlow, chen2017research}, work zone safety frameworks \cite{sabeti2021toward, fang2018detecting, cai2020context}, intelligent traffic monitoring \cite{Chen2020WITM}, and autonomous driving \cite{cui2019multimodal, hussain2022vision}. The industrial CPS (ICPS) and city-based CPS (CCPS)  use these prediction models for smart surveillance \cite{Isern2020Reconfig, wu2019deep, pazho2023ancilia}, human/pedestrian/worker safety \cite{deniz2022efficient, wang2022intelligent}, anomaly detection and prediction \cite{jones2014anomaly, pustokhina2021automated}, and path planning \cite{farooq2021flow}. 

Path prediction algorithms are vital to the decision-making process in TCPS, ICPS, and CCPS feedback systems. They are used in safety applications for predicting future actions and potential positions, autonomous vehicles for avoiding obstacles and defining a safe path and velocity, and anomaly detection to predict future anomalous behaviors from past trajectories. These algorithms are used at the local edge nodes for real-time detection and prediction \cite{Isern2020Reconfig, sanchez2021real, jeon2020scale}. However, as the path prediction architectures are often designed for specific domains and subjects in focus, they struggle to adapt to changes in the domain. In the following, we review existing path prediction approaches in three primary domains and their {challenges}: (1) Vehicle Bird's-eye View, (2) Pedestrian Bird's-eye View, and (3) Pedestrian High-angle View.

\begin{figure*}[!b]
        \centering
               \includegraphics[width=1\linewidth, trim={20px 200px 20px 200px}, clip]{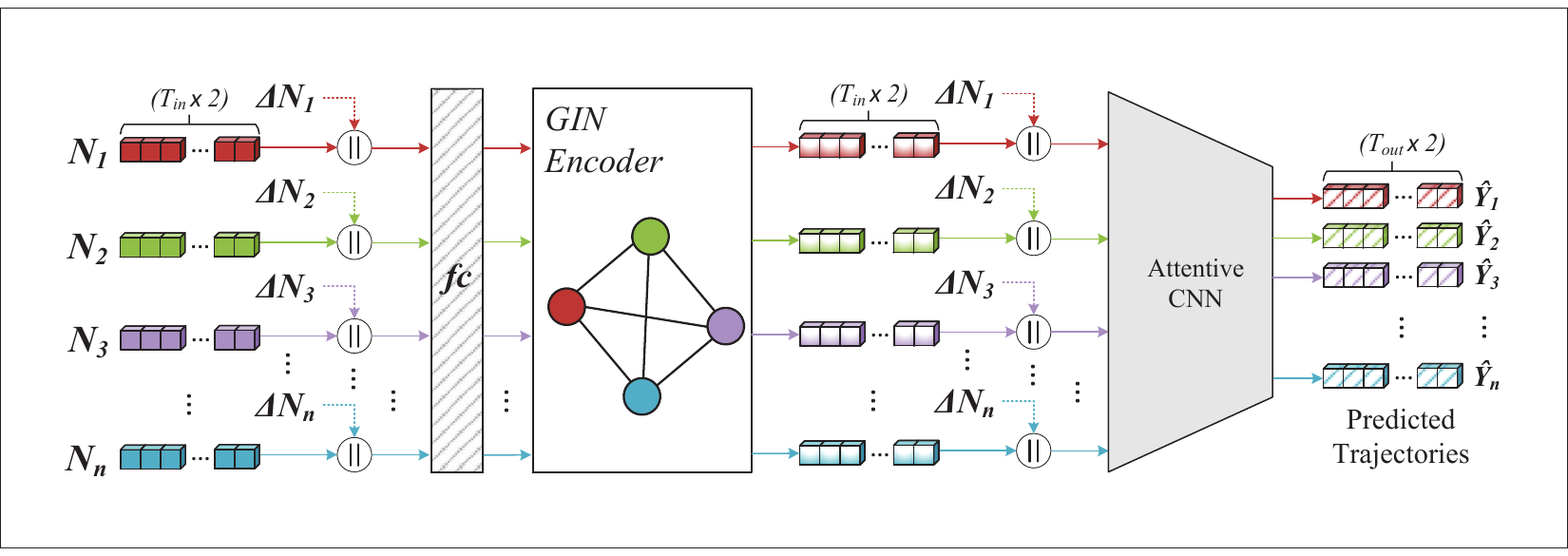}
                
                \caption{Pishgu formulation visualization. The input $N_i$ refers to a vector of size $T_{in}\times2$ for each node/subject where $T_{in}$ is the input window size. $\Delta N_i$ is the relative vector for each node. $fc$ refers to a fully-connected layer. The output is a $n\times T_{out}\times 2$ vector, with $T_{out}$ being the output window size. Best seen in color.}
                                
                \label{fig:pishgu}
                \vspace{-10pt}
\end{figure*}

\subsection{Vehicle Bird's-eye View Path Prediction}
The bird's-eye view is utilized by TCPS \cite{chwa2018closing, katariya2021deeptrack} and ICPS \cite{lian2020cyber} frameworks to generate a local map of the environment to comprehend the interdependencies of subjects moving at high speeds. Many approaches focus on predicting the future path of vehicles in a bird's-eye view setting on highways \cite{GSTCN2022, STALSTM2021, 2020gripplus}. These methods use real-world coordinates and measure the error in meters. CS-LSTM \cite{CSlstm18}, a pioneering paper in vehicle path prediction, used Long Short-term Memory (LSTM) encoder-decoder model with convolutional social pooling. GRIP++ \cite{2020gripplus}, an extension to \cite{grip2019}, uses fixed and dynamic graphs with an LSTM encoder-decoder model to grasp the surrounding dynamics and predict the trajectories of the vehicles in a scene. STA-LSTM \cite{STALSTM2021} utilizes Spatio-temporal attention along with LSTM to predict a vehicles' future path and increase the explainability of the predictions. \cite{Pip2021} incorporates the ego vehicle's planned path into the prediction of the surrounding vehicles' future paths to enhance predictions in an autonomous vehicle setting. \cite{GSAN} uses a graph self-attention network to understand both spatial and temporal interactions among the many vehicles in a scene and is used for both path prediction and lane-changing classification. In \cite{HEAT}, a three-channel framework with a heterogeneous edge-enhanced graph attention network is proposed to deal with the inherent heterogeneity of the different vehicles in a given scene. DeepTrack \cite{katariya2021deeptrack} introduces temporal and depthwise convolutions to provide a more robust encoding of vehicle dynamics while reducing computation and parameters, resulting in a faster, lighter-weight network with competitive accuracy. \cite{GSTCN2022} proposed recently, utilizes a graph-based spatial-temporal convolutional network and a gated recurrent unit to predict future vehicle paths. They also propose a weighted adjacency matrix for understanding the intensity of influence between different vehicles.

\subsection{Pedestrian Bird's-eye View Path Prediction}
Like vehicle path prediction, pedestrian path prediction often relies on a bird's-eye view of real-world coordinates and measures the error in meters. Many works have tackled the problem of path prediction in recent years. Most of them only focus on multi-future path prediction \cite{yue2022human, zhou2021sliding, mangalam2021goals,wong2021view} (k=5, 8, 15, 20, ..., where k shows the number of predicted trajectories for each subject), and evaluate the model by picking the best out of several predicted paths. However, generating multiple outputs per subject in real-time scenarios is not helpful in real-world applications. Therefore, we primarily focus on the works that perform single future path prediction analysis. \cite{alahi2016social} uses LSTM modules to predict the trajectories of all pedestrians in a scene jointly. Several works have used attention mechanisms to solve path prediction problems better. \cite{vemula2017social} introduces attention to model the importance of social interactions without relying on proximity. \cite{zhao2020spatial} also leverages a spatio-temporal attention module to learn which social interaction has a more critical role in predicting the pedestrian's path. Trajectron++ \cite{salzmann2020trajectron++}, designed for integration with robotic frameworks, utilizes a graph-structured recurrent model and heterogeneous data. While the focus is on pedestrians, Trajectron++ also predicts the future paths of vehicles to understand better how pedestrians might react to them. \cite{mendieta2021carpe} utilizes graph isomorphism networks and a lightweight Convolutional Neural Network (CNN) for path prediction, considerably reducing computation and model size and targeting real-time applications. SSAGCN \cite{lv2021ssagcn} uses an attention graph convolutional network and defines a new formulation to consider both social interactions and environmental factors as they can change the path pedestrians may choose.

\subsection{Pedestrian High-angle View Path Prediction}
Applications that require pedestrian path prediction only sometimes have access to bird's-eye view cameras or real-world coordinates. It is logical to use high-angle views and pixel space $(x,y)$ coordinates in addition to the traditional bird's-eye view models. This is common for CCPS \cite{Isern2020Reconfig} and ICPS \cite{deniz2022efficient, Chen2021TowardsOH} as the cameras are often mounted at the top of buildings or traffic lights and on the walls in respective systems. \cite{li2022graph} integrates multiple graph-based spatial transforms and trajectory smoothing to exploit temporal information and correct temporally inconsistent trajectories. SimAug \cite{liang2020simaug} utilizes synthetic data to improve the robustness of learned representations with the goal of better generalization in unseen contexts. Peeking into the Future \cite{liang2019peeking} proposes a multi-task model to predict both future paths and future activities, with the belief that understanding the future activity is highly informative to the future path. Multiverse \cite{liang2020garden} builds upon \cite{liang2019peeking} by introducing synthetic data and multi-scale location information and replacing graphs with Recurrent Neural Networks (RNNs). ScePT \cite{chen2022scept} proposes predicting paths at the scale of "cliques" rather than for each individual. Similar to Trajectron++ \cite{salzmann2020trajectron++}, and GAIN \cite{liu2022multi}, ScePT predicts vehicle and pedestrian trajectories.

Some of the works have tried to tackle the problem of path prediction in more than one domain \cite{liu2022multi, salzmann2020trajectron++, liang2019peeking, zhao2020spatial}. However, they prefer not to evaluate their model on all three discussed domains. On the other hand, although path prediction has many CPS applications, more work needs to be done to benchmark the path prediction architectures' performance on edge devices. Works such as \cite{Chen_2021_ICCV, Shafiee_2021_CVPR, Wang_2021_WACV, peng2021sra} report the inference time of the models on sophisticated and powerful hardware resources such as GTX 2080Ti GPU, NVIDIA Quadro RTX 6000 GPU, and Intel Core i9-9880H Processor. To this end, it is not feasible for CPS edge-based applications to use such a framework, as edge platforms have limited power and system resources. Thus, previous approaches fail to generalize across several different domains and adapt to the stringent requirements of the CPS applications.

\section{Pishgu} \label{sec:pishgu}
Path prediction in different domains inherently depends upon the position, past movements, and end goals of the subjects present in a scene. The end goals and movement patterns vary significantly between different environments (highway, sidewalk, etc.) and subjects (vehicle, pedestrian). However, graph neural networks assist in understanding the varying patterns in the respective domains. Pishgu uses GIN to grasp the interdependencies of all subjects in a scene. Attention-based convolutions are utilized to highlight the important interdependencies and predict the future paths of all the subjects jointly. As we focus on real-time applications of our approach, Pishgu is designed to predict a single path ($K=1$) for each subject present in the scene. The overall structure of Pishgu can be seen in Figure \ref{fig:pishgu}.

\subsection{Problem Formulation}
Our goal is to predict the future paths of all the subjects in a scene, regardless of the domain of prediction. Keeping that in mind, we use both absolute and relative coordinates as inputs to Pishgu, defined as follows:

\begin{equation}
     \label{eq:input}
    \boldsymbol{N_i}=[N_{i}^{t_{-T_{in}}} ,N_{i}^{t_{1-T_{in}}} ,\cdots , N_{i}^{t_{0}}]
\end{equation}    
\begin{equation}
\label{eq:rel_input}
    \boldsymbol{\Delta N_i} = [\Delta N_{i}^{t_{-T_{in}}} ,\Delta N_{i}^{t_{1-T_{in}}} \cdots  ,\Delta N_{i}^{t_{0}}]
\end{equation}        
    
where $N_{i}^{t}=\left<x_{i}^{t}, y_{i}^{t}\right>$ is the position of subject $i$ (vehicle or pedestrian)  at time $t$ and $T_{in}$ is the number of time steps that the model observes for prediction. $\Delta N_{i}^{t} = N_i^t - N_i^{t_{-T_{in}}}$ in equation \ref{eq:rel_input} represents the relative coordinates of subject $i$ in time $t$ with respect to the location of the subject at time $-T_{in}$. This approach is specifically adopted for exploiting the features of the graph-based architecture. The model is aware of the surrounding environment which plays a part in the path prediction process similar to humans' decision-making process while driving and walking. 

As mentioned previously, Pishgu predicts a single path for each subject at every time step, and the outputs of the model are formulated as:
\begin{equation}
\label{eq:output}
    \boldsymbol{\hat{Y_{i}}}=\left[
        \hat{Y_{i}}^{t_{1}},\hat{Y_{i}}^{t_{2}}, \cdots, \hat{Y_{i}}^{T_{out}}
    \right],
\end{equation}
where $ \hat{Y_{i}}^{t}$ represents the predicted trajectory of subject $i$ at time t in the future up to $T_{out}$ time steps.

\subsection{Architecture}

Pishgu's architecture (see Figure \ref{fig:pishgu}) is relatively simple compared to many modern path prediction models. This simplicity in our design is intentional; with CPS applications in mind, optimizing the number of parameters can improve performance in real-world scenarios. After the calculation of input features, $\boldsymbol{N}$ and $\boldsymbol{\Delta N}$ are concatenated and passed through a single fully-connected layer $fc$ as shown in Figure \ref{fig:pishgu}. In the next step, Pishgu leverages a Graph Neural Network (GNN) for embedding the input features. There has been a surge in curiosity towards Graph Neural Networks in recent years because of their power to represent complex interactions, and non-Euclidean data \cite{xu2018powerful,velivckovic2017graph, kipf2016semi}. Many different approaches have been proposed with neighbor aggregation and message-passing methods. The final goal of GNNs is to construct a maximally discriminative representation. This means that two nodes are mapped to the same location in the embedding space only if they are identical. \cite{xu2018powerful} came up with a simple yet powerful new formulation of GNNs, namely GIN that is shown to be as powerful as the Weisfeiler-Lehman test \cite{leman1968reduction} which is a test that answers the question of whether two graphs are identical or not. We draw motivation from the work of \cite{xu2018powerful} and adapt the network to our specific requirements. Our model constructs a fully connected graph $\mathcal{G}$ = ($\mathcal{V}, \mathcal{E}$) where the nodes ($\mathcal{V}$) are the subjects of interest present in the frame, and the edges ($\mathcal{E}$) represent their interactions. The fully connected structure assures that all the possible interactions between subjects are considered, and the network can extract all the important information from other neighbors. Pishgu utilizes a modified version of the aggregation function introduced by \cite{xu2018powerful} and constructs $f_{i}^{\prime}$ (the aggregated feature for node $i$) as follows:
\begin{equation}
f_{i}^{\prime}=MLP_{0}\left((1+\theta) \cdot f_{i}\right)+ MLP_{1}\left(\sum_{j\in \mathcal{V}(i)} f_{j}\right)
\label{eq:GIN}
\end{equation}

where $i$ is $i^{th}$ subject in the scene, $MLP_{0}$ and $MLP_{1}$ are Multi-layer Perceptron (MLP) operators each with a single hidden layer, $\mathcal{V}(i)$ is the set of node $i$ neighbors and $\theta$ is a trainable parameter. $MLP_{0}$  and $MLP_{1}$ are applied to the features of node $i$ and the aggregation of the features from neighbor nodes, respectively. Having two separate MLPs improves the network's ability to extract richer features and better integrate neighboring nodes in the context \cite{mendieta2021carpe}. Keeping the real-time performance in mind, a single graph operation is performed across the entire graph, which is enough since the graph is fully connected and all the features can be propagated in one step. Before the final task of path prediction, the output vector of the GIN block (equation \ref{eq:GIN}) for each subject is concatenated with the respective relative coordinate (equation \ref{eq:rel_input}).

Pishgu makes use of an attentive convolutional neural network for the final path predictions. Studies have shown that adding attention to CNNs can improve their representation power and help them generalize better \cite{cbam18}. The attention enables the convolutional predictor to dynamically decide the importance of embedded features generated by the GIN Encoder in the previous step and how much of this information should be used in the final prediction. Keeping the computational complexity in mind for CPS applications, the predictor structure consists of only seven layers, with three layers of 2D convolution each followed by a layer of attention module \cite{cbam18} and a final $1\times1$ convolutional layer for forming the predicted trajectories. The first convolutional layer uses a $2 \times 2$ kernel, and the subsequent two convolutional layers use a $2\times1$ kernel size. The architecture is designed to capture the fast and slow movements of the subjects. The attention module works based on two pillars: channel attention and spatial attention. Channel attention tries to perceive what is essential in the input feature map \cite{NIU202148}. To accomplish this task, the channel attention module first performs average pooling and max pooling on the input feature map to encapsulate the input features. These pooling operations improve the module's efficiency by reducing the feature map size.

In the next step, pooled features are fed to an MLP with one hidden layer to create the channel attention map. The formulation of the channel attention module can be summarized as:
\begin{equation}
\begin{aligned}
\mathbf{AT_{c}}(\mathbf{F}) = {} & MLP_{2} (Pool_{avg}(\mathbf{F})) \\
                                & +MLP_{2} (Pool_{max}(\mathbf{F}))
\end{aligned}
\end{equation}
Where $\mathbf{F}$ is the input feature map, $MLP_2$ is an MLP with one hidden layer, and $\mathbf{AT}_{c}$ is the channel attention module. 
On the other hand, the goal of the spatial attention module is to locate the important features in the input \cite{NIU202148}. For efficiency, average pooling and max pooling are used again, but this time in the channel axis. The pooled features are concatenated and passed to a convolutional layer for generating the spatial attention map. In summary, the spatial attention module works based on the following equation:
\begin{equation}
\begin{aligned}
&\mathbf{AT_s}(\mathbf{F}) \\
&=\sigma\left(Conv^{7 \times 7}[(Pool_{avg}(\mathbf{F})) ; Pool_{max}(\mathbf{F})])\right) 
&
\end{aligned}
\end{equation}

where $\mathbf{F}$ is the input feature map, $\sigma$ is the sigmoid activation function, $Conv^{7 \times 7}$ is the convolutional layer with a kernel size of 7, and $\mathbf{AT}_{c}$ is the spatial attention module. 

Incorporating channel attention and spatial attention sequentially has shown significant improvements in the performance of CNNs \cite{woo2018cbam}. Thus, between each convolutional layer, there is an additional module consisting of the channel and spatial attention to improve the predictor's performance while keeping the network lightweight.  



\section{Domain Analysis}
Challenges and requirements found in path prediction are often domain-specific. The interactions and behaviors vary drastically from predicting a vehicle's path on a highway with hundreds of vehicles in a scene to predicting the path of a single pedestrian walking on a sidewalk with less than a dozen other people. A universal model should be able to adapt across these domains. Thus, understanding each domain's characteristics is vital for designing a universal path prediction architecture. Table \ref{tab:stat} highlights the complexities of various domains by presenting the number of subjects and samples available from widely adopted datasets in the respective domains.

\begin{table}[!htbp]
\renewcommand{\arraystretch}{1.2}
\centering
\caption{The statistics of four datasets used for path prediction. These parameters are reported after doing the conventional preprocessing steps discussed in section \ref{sec:domain}. $V_B$, $P_B$, and $P_H$ refer to Vehicle Bird's-eye view, Pedestrian Bird's-eye view, and Pedestrian High-angle view respectively. FPS is Frames Per Second.}
\vspace{-10pt}
\label{tab:stat}
\begin{adjustbox}{max width=1\columnwidth,center}
\begin{tabular}{c||c|c|c|c}
\rowcolor{DarkGray}
 & \begin{tabular}[c]{@{}c@{}} NGSIM \\ \cite{ NGSIM_i80, NGSIM_US101} \end{tabular} & \begin{tabular}[c]{@{}c@{}} ETH \\ \cite{pellegrini2009you} \end{tabular} & \begin{tabular}[c]{@{}c@{}} UCY \\ \cite{lerner2007crowds} \end{tabular} & \begin{tabular}[c]{@{}c@{}}ActEv/\\ VIRAT \cite{awad2018benchmarking}\end{tabular} \\ \hline \hline
Domain & $V_B$ & $P_B$ & $P_B$ & $P_H$ \\ \hline
\#Subjects & 19,698 & 749 & 1,456 & 1,059 \\ \hline
\#Samples & 13.3M & 12,035 & 62,393 & 130,289 \\ \hline
\#Frames & 108,033 & 2,044 & 4,397 & 41,199 \\ \hline
FPS & 5 & 2.5 & 2.5 & 2.5 
\end{tabular}
\end{adjustbox}
\vspace{-10pt}
\end{table}
\label{sec:domain}

\subsection{Vehicle Bird's-eye View}

Most deep learning architectures used for predicting vehicle paths in a highway environment \cite{CSlstm18, highway2022, congestion2021} use NGSIM datasets \cite{NGSIM_i80, NGSIM_US101}. NGSIM provides complex real-world scenarios and driver behaviors for various traffic patterns. In this paper, we use US-101 \cite{NGSIM_US101} and I-80 \cite{NGSIM_i80} from NGSIM, each with millions of data samples in bird's-eye view. Using a bird's-eye view for path prediction is natural as it provides an overall perspective of the surrounding environment. NGSIM data is collected from cameras mounted on the buildings around the freeways. However, the final dataset provides the data in bird's-eye format. This conversion to a bird's-eye view helps deep learning models learn complicated driver behaviors and their reactions to the surroundings. In this evaluation we exclude the trajectories of vehicles going for exits and merging lanes for both \cite{NGSIM_i80} and \cite{NGSIM_US101}.

\subsection{Pedestrian Bird's-eye View}
Two widely used pedestrian path prediction datasets are ETH \cite{pellegrini2009you} and UCY \cite{lerner2007crowds}. They consist of bird's-eye view annotations of several crowded scenes with complicated nonlinear pedestrian paths. The position of the pedestrians in these datasets is gathered in real-world coordinates in meters. The data points generally used for training are sampled at a rate of 2.5 Frames Per Second (FPS). ETH looks at two different scenes, ETH and HOTEL, whereas UCY looks at three different scenes, UNIV, ZARA1, and ZARA2. These datasets are especially useful for training models that focus on drone-related applications or environment monitoring. They do not include any data for other points of view such as high-angle or side views.

\begin{figure}[h]
\vspace{-10pt}
\begin{tikzpicture}
  \begin{axis}
    [
    xmajorgrids=true,
    width= \linewidth,
    trim left = 1cm,
    ytick={1,2,3,4},
    xlabel = Number of Samples,
    yticklabels={NGSIM, ETH , UCY, \begin{tabular}[c]{@{}c@{}}ActEv/\\ VIRAT  \end{tabular}},
    xticklabels={0, 1, 3, 10, 32, 100, 315},
    height = 0.7\linewidth,
    ]
    \addplot+[
    boxplot prepared={
      median=2.1,
      upper quartile=2.2,
      lower quartile=2,
      upper whisker=2.4,
      lower whisker=0
    }, draw=black, fill=Green,
    ] coordinates {};
    \addplot+[
    boxplot prepared={
      median=0.7,
      upper quartile=0.9,
      lower quartile=0.5,
      upper whisker=1.4,
      lower whisker=0
    }, draw=black, fill=Purple,
    ] coordinates {};
    \addplot+[
    boxplot prepared={
      median=0.9,
      upper quartile=1.2,
      lower quartile=0.6,
      upper whisker=1.9,
      lower whisker=0
    }, draw=black, fill=LightBlue,
    ] coordinates {};
    \addplot+[
    boxplot prepared={
      median=0.5,
      upper quartile=0.6,
      lower quartile=0.3,
      upper whisker=1.1,
      lower whisker=0
    }, draw=black, fill=LightRed,
    ] coordinates {};
  \end{axis}
\end{tikzpicture}
\vspace{-10pt}
\caption{Distribution of number of samples in a frame for ActEv/VIRAT \cite{awad2018benchmarking}, UCY \cite{lerner2007crowds}, ETH \cite{pellegrini2009you} and NGSIM \cite{NGSIM_i80, NGSIM_US101} datasets. X-axis shows the number of samples in the frame. Please note that the box and whiskers are drawn in logarithmic space for better visualization, and the number of samples per frame can be translated to the number of unique subjects in one frame.}
\label{fig:data_dist}
\vspace{-15pt}
\end{figure}
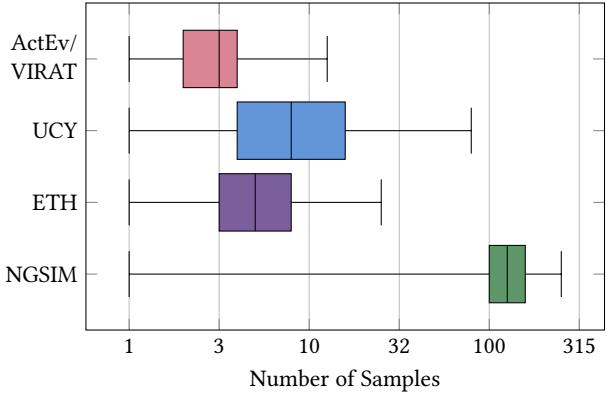

\subsection{Pedestrian High-angle View}

\begin{table*}[!b]
\renewcommand{\arraystretch}{1.3}
\centering
\caption{Vehicle path prediction error and complexity comparison of Pishgu with contemporary approaches. Best is in bold text. Second best is
underlined.}
\label{tab:ngsim_results}%
\vspace{-10pt}
  \begin{adjustbox}{max width = \linewidth,center}
    \begin{tabular}{c||c|c|c|c|c||c|c||c}
    \rowcolor{DarkGray} {} & \multicolumn{5}{c||}{RMSE (m)}        &  &  &  \\  \cline{2-9} \rowcolor{Gray}
          & 1s    & 2s    & 3s    & 4s    & 5s    &   {ADE (m)}    &   {FDE (m)}    & \multicolumn{1}{c}{Params (K)}  \\ \hline \hline
    
    CS-LSTM \cite{CSlstm18} & 0.63  & 1.27  & 2.09  & 3.10   & 4.37 & 2.29  & 3.34   & \multicolumn{1}{c}{191} \\ \hline
    DeepTrack \cite{katariya2021deeptrack} & 0.47  & 1.08  & 1.83  & 2.75  & 3.89  & 2.01  & 3.25  & \multicolumn{1}{c}{\underline{109}}  \\ \hline
    
    GRIP++ \cite{2020gripplus} &  0.38  &   0.89 &    {1.45}   & 2.14       &    \underline{2.94}   &    1.61       &    -   & \multicolumn{1}{c}{-}  \\ \hline
    Pip \cite{Pip2021}  & 0.55  & 1.18  & 1.94  & 2.88  & 4.04  &   2.18    &    -   & \multicolumn{1}{c}{-}   \\ \hline
    GSTCN \cite{GSTCN2022}& 0.44  & \underline{0.83}  & \underline{1.33}   & \underline{2.01}  & 2.98  &   \underline{1.52 }   &   -    & \multicolumn{1}{c}{\textbf{49.8}}   \\ \hline
    STA-LSTM \cite{STALSTM2021} & \underline{0.37}  & 0.98  & 1.71  & 2.63  & 3.78  & 1.89  & \underline{3.16}  & \multicolumn{1}{c}{124} \\ \hline
    \rowcolor{LightGray} \textbf{Pishgu (Ours)} & \textbf{0.15}  & \textbf{0.46}  & \textbf{0.82}  & \textbf{1.25}  & \textbf{1.74}  & \textbf{0.88}  & \textbf{1.96}  & \multicolumn{1}{c}{132}  \\
    \end{tabular}%
\end{adjustbox}
\vspace{-10pt}
\end{table*}%

It is unrealistic to assume that a bird's-eye view angle is always available. Most surveillance cameras are placed at a high location (e.g. the side of a building, on a light pole) to have an overview of the scene from a high-angle view. Thus, path prediction models in real-world scenarios should be able to work with different camera views and angles specially if they want to cover normal video surveillance setups. However, most of the works in the path prediction field focus on the bird's-eye view. On the other hand, in end-to-end real-world environment monitoring and surveillance systems, real-world coordinates are unavailable and the locations of subjects are represented in pixel space. This disconnect between real-world scenarios and most existing algorithms makes many proposed methods ill-suited for realistic path prediction. To address this issue, works such as \cite{liang2019peeking, liang2020garden, liang2020simaug, li2022graph} have used ActEv/VIRAT \cite{awad2018benchmarking} for path prediction. Since then, it has become a standard benchmark for path prediction in pixel space and high-angle view. The primary use of ActEV/VIRAT is activity recognition in challenging real-world scenarios in the surveillance domain. However, the diversity of camera angles and locations in this dataset is advantageous for real-world path prediction as well. Following the same preprocessing steps as Next \cite{liang2019peeking}, we downsample the frame rate to 2.5 FPS. The center of the bounding box for a person is used for the extraction of the subject's location.

Looking at Table \ref{tab:stat}, we can see that in the human bird's-eye view domain, the two publicly available datasets ETH \cite{pellegrini2009you} and UCY \cite{lerner2007crowds} are relatively small compared to datasets available in other domains. NGSIM \cite{NGSIM_i80, NGSIM_US101} has the most number of samples by far and is a much more crowded dataset in terms of the number of samples in each frame. Figure \ref{fig:data_dist} clearly shows that the number of samples in the frame for ETH and UCY datasets is more than ActEv/VIRAT \cite{awad2018benchmarking}, but still much less than the number of samples in NGSIM dataset which even goes up to 248 vehicle in a single frame. Due to these distinctive characteristics that each domain has, a general path prediction method should be able to adapt to each domain to be able to satisfy each domain's requirements.

\section{Experiments} \label{sec:Experiments}

This section evaluates our path prediction method in all three domains discussed in Section \ref{sec:domain}. We compare Pishgu with SotA models that report their performance in the respective domains in three different error measurements and model sizes. All of the evaluations in this article have been conducted on a server containing an NVIDIA Tesla V100 GPU, 2x AMD EPYC 7513 CPUs, and 256GB of RAM. Additionally, to prove the usability of our architecture for CPS applications, latency and throughput through real-time inference are reported on NVIDIA Jetson Xavier NX and Jetson Nano embedded platforms. The following sub-sections discuss the definitions of each error measurement and the performance of Pishgu against current path prediction approaches in all the domains.

\subsection{Evaluation Metrics}

\begin{table*}[!t] 
\caption{Path prediction error comparison of Pishgu in pedestrian bird's-eye view domain with contemporary approaches. ADE and FDE are reported in meters. Best is in bold text. Second best is
underlined.}
\vspace{-8pt}
\label{tab:human_bird}
\renewcommand{\arraystretch}{1.3}
\centering
\begin{adjustbox}{max width = \linewidth,center}
\begin{tabular}{c||cc||cc||cc||cc||cc||cc||c}
\rowcolor{DarkGray}
                                  & \multicolumn{2}{|c||}{ETH}                  & \multicolumn{2}{c||}{HOTEL}                & \multicolumn{2}{c||}{UNIV}                 & \multicolumn{2}{c||}{ZARA1}                & \multicolumn{2}{c||}{ZARA2}                & \multicolumn{2}{c||}{AVG}      &             \\ \cline{2-14}
                                  \rowcolor{Gray}
                                  & \multicolumn{1}{|c|}{ADE}  & FDE  & \multicolumn{1}{c|}{ADE}  & FDE  & \multicolumn{1}{c|}{ADE}  & FDE  & \multicolumn{1}{c|}{ADE}  & FDE  & \multicolumn{1}{c|}{ADE}  & FDE  & \multicolumn{1}{c|}{ADE}  & FDE & Params (M) \\ \hline \hline
\multicolumn{1}{c||}{Linear}       & \multicolumn{1}{c|}{1.33}          & 2.94          & \multicolumn{1}{c|}{0.39}          & 0.72          & \multicolumn{1}{c|}{0.82}          & 1.59          & \multicolumn{1}{c|}{0.62}          & 1.21          & \multicolumn{1}{c|}{0.77}          & 1.48          & \multicolumn{1}{c|}{0.79}          & 1.59     & -     \\ \hline
\multicolumn{1}{c||}{LSTM}         & \multicolumn{1}{c|}{1.09}          & 2.41          & \multicolumn{1}{c|}{0.86}          & 1.91          & \multicolumn{1}{c|}{0.61}          & 1.31          & \multicolumn{1}{c|}{0.41}          & 0.88          & \multicolumn{1}{c|}{0.52}          & 1.11          & \multicolumn{1}{c|}{0.70}           & 1.52     & -     \\ \hline
\multicolumn{1}{c||}{Social-LSTM \cite{alahi2016social}} & \multicolumn{1}{c|}{1.09}          & 2.35          & \multicolumn{1}{c|}{0.79}          & 1.76          & \multicolumn{1}{c|}{0.67}          & 1.40           & \multicolumn{1}{c|}{0.47}          & 1.00             & \multicolumn{1}{c|}{0.56}          & 1.17          & \multicolumn{1}{c|}{0.72}          & 1.54     & -     \\ \hline
\multicolumn{1}{c||}{Next \cite{liang2019peeking}}         & \multicolumn{1}{c|}{0.88}          & 1.98          & \multicolumn{1}{c|}{{0.36}}          & {0.74}          & \multicolumn{1}{c|}{0.62}          & 1.32          & \multicolumn{1}{c|}{0.42}          & 0.90           & \multicolumn{1}{c|}{0.34}          & 0.75          & \multicolumn{1}{c|}{{0.52}}          & 1.14   & 3.95       \\ \hline
\multicolumn{1}{c||}{ST-Attention \cite{zhao2020spatial}} & \multicolumn{1}{c|}{{{0.85}}} & {1.85}     & \multicolumn{1}{c|}{{0.32}} & {{0.66}} & \multicolumn{1}{c|}{{0.63}} & {1.33} & \multicolumn{1}{c|}{{0.42}} & {0.91} & \multicolumn{1}{c|}{{0.34}} & {0.73} & \multicolumn{1}{c|}{{0.51}} & 1.10  & 1.98\\ \hline
\multicolumn{1}{c||}{CARPe \cite{mendieta2021carpe}}        & \multicolumn{1}{c|}{0.80}           & {1.48} & \multicolumn{1}{c|}{0.52}          & 1.00             & \multicolumn{1}{c|}{{0.61}}          &{ 1.23 }         & \multicolumn{1}{c|}{{0.42}}          & {0.84}          & \multicolumn{1}{c|}{{0.34}}          & {0.69}          & \multicolumn{1}{c|}{0.54}          & 1.05   & \textbf{0.10}       \\ \hline
\multicolumn{1}{c||}{Trajectron++ \cite{salzmann2020trajectron++}} & \multicolumn{1}{c|}{{0.71}} & {1.66}          & \multicolumn{1}{c|}{\underline{0.22}} & \underline{0.46} & \multicolumn{1}{c|}{{0.44}} & \underline{1.17} & \multicolumn{1}{c|}{{0.30}} & \underline{0.79} & \multicolumn{1}{c|}{{0.23}} & \underline{0.59} & \multicolumn{1}{c|}{{0.38}} & \underline{{0.93}} & - \\ \hline
\multicolumn{1}{c||}{ScePT \cite{chen2022scept}} & \multicolumn{1}{c|}{\textbf{0.19}} & \underline{1.33}          & \multicolumn{1}{c|}{\textbf{0.18}} & {1.12} & \multicolumn{1}{c|}{\textbf{0.19}} & {1.19} & \multicolumn{1}{c|}{\textbf{0.18}} & {1.10} & \multicolumn{1}{c|}{\underline{0.19}} & {1.20} & \multicolumn{1}{c|}{\textbf{0.19}} & {{1.19}} & - \\ \hline
\multicolumn{1}{c||}{SSAGCN \cite{lv2021ssagcn}} & \multicolumn{1}{c|}{{\underline{0.30}}} & \textbf{0.59}          & \multicolumn{1}{c|}{\underline{0.22}} & {\textbf{0.42}} & \multicolumn{1}{c|}{\underline{0.25}} & \textbf{0.47} & \multicolumn{1}{c|}{\underline{0.20}} & \textbf{0.39} & \multicolumn{1}{c|}{\textbf{0.14}} & \textbf{0.28} & \multicolumn{1}{c|}{\underline{0.22}} & {\textbf{0.43}}  & -\\ \hline
\rowcolor{LightGray}
\multicolumn{1}{c||}{\textbf{Pishgu (Ours)}}       & \multicolumn{1}{c|}{1.10}           & 2.24          & \multicolumn{1}{c|}{1.17}          & 2.17          & \multicolumn{1}{c|}{0.67}          & 1.37          & \multicolumn{1}{c|}{0.45}          & 0.91          & \multicolumn{1}{c|}{0.36}          & 0.73          & \multicolumn{1}{c|}{0.75}          & 1.48 & \underline{0.11} \\ 

\end{tabular}
\end{adjustbox}
\vspace{-10pt}
\end{table*}

\textbf{Average Displacement Error (ADE): } The average $L_2$ distance between the predicted coordinates ($\hat{Y}$) and the ground truth coordinates ($Y$) over all $T_{out}$ predicted time steps and all subjects of interest (N) available in the scene:
\begin{equation}
\mathrm{ADE}=\frac{\sum_{i=1}^{N} \sum_{t=1}^{T_{out}}\left\|Y_{i}^{t}-\hat{Y}_{i}^{t}\right\|_{2}}{N * T_{out}}
\end{equation}

\textbf{Final Displacement Error (FDE): } The average $L_2$ distance between the predicted coordinates ($\hat{Y}$) and the ground truth coordinates ($Y$) of the last predicted time step over all subjects of interest (N) available in the scene:

\begin{equation}
\mathrm{FDE}=\frac{\sum_{i=1}^{N}\left\|Y_{i}^{T_{out}}-\hat{Y}_{i}^{T_{out}}\right\|_{2}}{N}
\end{equation}

\textbf{Root Mean Square Error (RMSE):}
The RMSE at time $t$ is the square root of the mean square error between the predicted path ($\hat{Y}$) and the ground truth path ($Y$) of the subjects of interest in the scene:
\begin{equation}
    RMSE^{t}=\sqrt{\frac{1}{N}\sum_{i=1}^{N}{(Y_{i}^{t}-\hat{Y}_{i}^{t})}^{2}}
\end{equation}
 
We report and compare the complexity of our model against the SotA approaches in terms of number parameters. In Section \ref{sec:realtime}, we calculate the latency of prediction per sample in milliseconds and the throughput in FPS.

\subsection{Vehicle Bird's-eye View}

Recent works \cite{GSTCN2022, aitp2022,dginet2022} have used the NGSIM public datasets for vehicle path prediction as it provides many complex highway scenarios. The dataset is split into a 70\% training set, 10\% validation set, and 20\% test set. Pishgu has been trained for 40 epochs with a learning rate of 0.01 and the Adam optimizer \cite{kingma2014adam}.

As shown in Table \ref{tab:ngsim_results}, we review the performance of Pishgu in terms of RMSE, ADE, FDE, and model size against best-in-class models \cite{CSlstm18, katariya2021deeptrack, 2020gripplus, Pip2021, GSTCN2022, STALSTM2021} in vehicle bird's-eye view domain. Reporting RMSE is crucial in vehicle path prediction as autonomous vehicles evaluate the future trajectories of surrounding subjects for every second up to a few seconds. Pishgu, like the other models, observes the input for 15-time steps (3 seconds) and predicts the path of all the vehicles in a scene for 25-time steps (5 seconds). Regarding RMSE, Pishgu performs better at all the time steps than the other models. It reduces the RMSE by 44\%, 38\%, and 37\% for 2$^{nd}$, 3$^{rd}$, and 4$^{th}$ sec as compared to \cite{GSTCN2022}. Compared to best-performing models for 1$^{st}$ and 5$^{th}$ sec, Pishgu reduces the RMSE by 50\% and 40\%. Similar performance is observed in ADE and FDE, where we improve the SotA error rates by 42\% and 61\%, respectively. As most of the frames in NGSIM are crowded, it helps the attention module grasp the most important relational behaviors between the subjects for path prediction. Thus, the attention module is the most effective for the NGSIM dataset.

In model size, Pishgu is comparable to most contemporary deep-learning models. However, \cite{GSTCN2022} reports less than half the parameters than Pishgu. The smaller size of GSTCN is due to using a simple CNN for path prediction and restricting the graph to two laterally adjacent lanes and $\pm$100 meters. In contrast, Pishgu uses attention-based CNN with a graph of the entire scene to grasp the overall environment provided by the NGSIM dataset.

\subsection{Pedestrian Bird's-eye View}

In the pedestrian bird's-eye view domain, we evaluated our model on ETH \cite{pellegrini2009you}, and UCY \cite{lerner2007crowds}, the characteristics of which were detailed in Section \ref{sec:domain}. We will adopt the same strategy for evaluation as previous works in this domain \cite{gupta2018social} and train our model with a leave-one-out approach on combined ETH and UCY datasets. As input, the model observes 8-time steps (3.2 seconds) and predicts the coordinates of the pedestrian for the future 12-time steps (4.8 seconds). The model has been trained for 80 epochs with a learning rate of $0.01$ and Adam optimizer \cite{kingma2014adam}.

In the field of path prediction, it is common for models to generate 20 outputs (k=20) for each subject in the scene and report the ADE and FDE based on the best of the 20 predicted paths for each pedestrian. However, this method is unsuitable for real-time scenarios, especially those deployed on embedded platforms with limited resources. Thus, we will compare our model only to the approaches that report their single-future path prediction evaluation in the pedestrian bird's-eye view domain. In our comparisons, we include a simple Linear regressor used for path prediction. Another baseline method we consider is LSTM, a simple LSTM encoder-decoder model used for path prediction. Table \ref{tab:human_bird} shows that Pishgu struggles with path prediction in this domain. This is due to the fact that ETH\cite{pellegrini2009you} and UCY \cite{lerner2007crowds} are relatively small datasets and can not provide a sufficient amount of training data for the convolution attention mechanism to be trained well, as discussed in Section \ref{sec:domain}. 

We also report the model complexity in terms of the number of parameters, but most previous works do not examine this crucial factor. Looking at Table \ref{tab:human_bird}, we can see that Pishgu has a competitive model size compared to \cite{mendieta2021carpe} and around $36\times $ smaller model size compared to Next \cite{liang2019peeking}.

\begin{table*}[!t]
\centering
\caption{Experiment results on ActEV/VIRAT \cite{awad2018benchmarking} dataset. The results are reported in pixel space. Best is in bold text. Second best is underlined.}
\vspace{-10pt}
\label{tab:pixel}
\renewcommand{\arraystretch}{1.3}
\centering
\begin{adjustbox}{max width = \linewidth,center}
\begin{tabular}{c||c|c|c|c|c|c|c|c|c}
\rowcolor{DarkGray}
 & \begin{tabular}[c]{@{}c@{}}Social-LSTM \\ \cite{alahi2016social} \end{tabular} & \begin{tabular}[c]{@{}c@{}}Social-GAN\\ (V) \cite{gupta2018social} \end{tabular} & \begin{tabular}[c]{@{}c@{}}Social-GAN \\ (PV) \cite{gupta2018social} \end{tabular} & \begin{tabular}[c]{@{}c@{}}Next \\ \cite{liang2019peeking} \end{tabular}  & \begin{tabular}[c]{@{}c@{}} Multiverse \\ \cite{liang2020garden} \end{tabular}  & \begin{tabular}[c]{@{}c@{}} SimAug \\ \cite{liang2020simaug} \end{tabular}   & \begin{tabular}[c]{@{}c@{}} ST-MR \\ \cite{li2022graph} \end{tabular} & \begin{tabular}[c]{@{}c@{}} ST-Attention \\ \cite{zhao2020spatial} \end{tabular} & \textbf{Pishgu (Ours)} \\ \hline \hline
ADE & 23.10 & 30.40 & 30.42 & 19.78 & {18.51} & 21.73 & 18.58 & \underline{18.39} &\cellcolor{Gray} \textbf{14.11} \\ \hline
FDE & 44.27 & 61.93 & 60.70 & 42.43 & \underline{35.84} & 42.22 & 36.08 & 38.11 & \cellcolor{Gray} \textbf{27.96} \\ \hline
Params (M) & - & - & - & {3.95} & - & - & - & \underline{1.98} &\cellcolor{Gray} \textbf{0.11}
\end{tabular}
\end{adjustbox}
\vspace{-10pt}
\end{table*}

\begin{table*}[b!]
\centering
\caption{Latency and throughput evaluation on Jetson Xavier NX CPU and  Jetson Nano CPU in milliseconds and FPS, respectively. Samples per Frame is an average. FPS is Frames Per Second, and $V_B$, $P_B$, and $P_H$ refer to Vehicle Bird's-eye view, Pedestrian Bird's-eye view, and Pedestrian High-angle view, respectively.}
\vspace{-10pt}
\label{tab:latency}
\renewcommand{\arraystretch}{1.3}
\begin{adjustbox}{max width = \linewidth,center}
\begin{tabular}{c||c|c||c|c||c|c}
    \rowcolor{DarkGray}
    \cellcolor{Gray} & \cellcolor{Gray} & \cellcolor{Gray} Samples & \multicolumn{2}{c||}{\cellcolor{DarkGray}  Jetson Xavier NX}  & \multicolumn{2}{c}{\cellcolor{DarkGray}  Jetson Nano} \\ \cline{4-7} 
    \rowcolor{Gray}
     & Domain & per   & Latency & FPS & Latency & FPS \\
    \rowcolor{Gray}
     &        & Frame & (ms)    &     & (ms)    &     \\ \hline \hline
NGSIM \cite{NGSIM_i80, NGSIM_US101}     & $V_B$  & 158 & 2.69  & 2.35   & 3.50  & 1.81   \\ \hline
UCY \cite{pellegrini2009you}            & $P_B$  & 8   & 1.44  & 86.81  & 2.41  & 51.81  \\ \hline
ETH \cite{lerner2007crowds}             & $P_B$  & 3   & 1.75  & 190.48 & 2.06  & 161.81 \\ \hline
ActEv/                                  & \multirow{2}{*}{$P_H$}  & \multirow{2}{*}{5}   & \multirow{2}{*}{2.83}  & \multirow{2}{*}{70.67}  & \multirow{2}{*}{3.40}   & \multirow{2}{*}{58.82}  \\
VIRAT \cite{awad2018benchmarking}       &        &     &       &        &       &        \\
\end{tabular}
\end{adjustbox}
\vspace{-10pt}
\end{table*}

\subsection{Pedestrian High-angle View}

Following the previous works \cite{liang2019peeking, liang2020garden, liang2020simaug, li2022graph}, we analyze the efficacy of Pishgu by training and testing it on the ActEV/VIRAT \cite{awad2018benchmarking} challenge dataset discussed in Section \ref{sec:domain}. We report ADE and FDE for evaluating the displacement error. The official training and validation sets have been used for collecting the mentioned evaluation results. As input, the model observes 8-time steps, or 3.2 seconds and predicts the path for the future 12-time steps or 4.8 seconds.

We compare our model to other models that make single-future path predictions in this domain since having multiple predicted paths per pedestrian in real-time applications is unrealistic. Results reported in pixel space can be seen in Table \ref{tab:pixel}. Social-LSTM \cite{alahi2016social}, and Social-GAN (both V and PV variants) results are based on the tests performed by \cite{liang2019peeking}. Comparison with other approaches makes evident the advantage of Pishgu with a $23.6\%$ and $22.6\%$ decrease in ADE and FDE with respect to the previous SotA model. These results push Pishgu to the top position in single-future path prediction by a large margin and show the benefit of an attentive CNN predictor in grasping the meaningful features generated by the GIN in the embedding stage.

Regarding the number of parameters, we can only compare Pishgu to Next \cite{liang2019peeking} and ST-Attention \cite{zhao2020spatial} since other works do not mention their model sizes. Pishgu, with around $18\times$ smaller model size compared to ST-Attention and the outstanding ADE and FDE results, is much more suitable for real-time deployment than previous works. 

\subsection{Real-time Evaluation} \label{sec:realtime}

Real-time path prediction applications are often implemented on resource-constrained embedded devices. Limited memory, limited power, and real-time implementation make lightweight, low-latency models necessary. We report the latency performance of Pishgu on multiple embedded platforms: Nvidia Jetson Xavier NX with a 15W dual-core Nvidia Carmel processor with 8GB of RAM and Jetson Nano with the 10W quad-core ARM Cortex-A57 processor with 2GB of RAM. We report the performance of two embedded platforms to demonstrate the adaptability of Pishgu in different resource-constrained environments. Both platforms are utilized at their respective highest power capacities. All the results on embedded platforms are calculated for the batch size. We test Pishgu for all three domains and report the latency per subject in milliseconds (ms) and throughput in terms of FPS.

The number of samples per frame is utilized to calculate the throughput to demonstrate the performance of Pishgu in real-world, real-time applications. As shown in Table \ref{tab:latency}, high numbers of vehicles in each frame of the NGSIM datasets ensure dense graphs, and millions of data samples help attention mechanism in feature map refinement \cite{cbam18}. However, the number of path predictions for the NGSIM dataset using Pishgu goes to more than 200 vehicles for a single frame. This, in turn, affects the overall throughput of the model even with low latency per sample. As the number of samples per frame for all the other datasets is meager compared to NGSIM, the throughput for other datasets is considerably higher. 

The throughput of NGSIM for the Jetson Xavier NX platform is 2.35 FPS which is 29.83\% better than 1.81 FPS for the lower-power Jetson Nano platform. Similarly, for pedestrian tracking, the throughput for UCY \cite{pellegrini2009you}, ETH \cite{lerner2007crowds}, and ActEv/VIRAT \cite{awad2018benchmarking} on the Jetson Xavier NX is 67.55\%, 17.72\%, and 20.15\% better than that on Jetson Nano embedded platform respectively. The superior performance of the Jetson Xavier NX can be credited to a higher operating power capacity of 15W with a higher clock frequency of 1.9 GHz. On the other hand, Jetson Nano with ARM cortex operates at 10W with a frequency of 1.5 GHz. Higher power distribution among the dual-core of Carmel processors also plays a role in justifying its performance. The average latency per sample for a bird's-eye vehicle view on Jetson Xavier NX is 2.69 ms, which is 23\% better than that of the Jetson Nano platform. Similar trends are observed in all the latency improvements comparison between two embedded platforms in pedestrian bird's-eye and high-angle view datasets. Hence, we demonstrate that Pishgu can be utilized in real-world applications using embedded platforms with reasonable latency and throughput performance.

\section{Conclusion}\label{sec:conclusion}
This paper proposes Pishgu, a universal path prediction architecture that leverages graph isomorphism networks and attention mechanisms for real-time CPS applications. We evaluate the competency of our architecture in three domains and present extensive domain analysis with their effects on the performance of our architecture in terms of error and complexity. The advantage of the proposed architecture is that it can adapt to multiple domains with best-in-class performance when trained and inferred on datasets in three different domains without any changes in the architecture. This is beneficial for the CPS applications as single architecture can be used for different application domains just by adjusting the model weights. Moreover, Pishgu is designed to be integrated at the local nodes of the real-world edge-based applications in mind, performing in real-time on embedded platforms. Pishgu achieves SotA performance for path prediction in vehicle bird's-eye and pedestrian high-angle view domains on NGSIM \cite{NGSIM_i80, NGSIM_US101}, and ActEV/VIRAT \cite{awad2018benchmarking}, respectively, by a considerable margin. Although vehicle high-angle view domain path prediction has many CPS applications, such as traffic monitoring and safety, there needs to be more work done in this domain. As a next step, we plan to focus on the vehicle high-angle view domain for path prediction.
\section*{Acknowledgements}
This research is supported by the National Science Foundation (NSF) under Award Numbers 1831795 and 1932524.

\bibliographystyle{ACM-Reference-Format}
\bibliography{sample-base}


\begin{thebibliography}{74}


\ifx \showCODEN    \undefined \def \showCODEN     #1{\unskip}     \fi
\ifx \showDOI      \undefined \def \showDOI       #1{#1}\fi
\ifx \showISBNx    \undefined \def \showISBNx     #1{\unskip}     \fi
\ifx \showISBNxiii \undefined \def \showISBNxiii  #1{\unskip}     \fi
\ifx \showISSN     \undefined \def \showISSN      #1{\unskip}     \fi
\ifx \showLCCN     \undefined \def \showLCCN      #1{\unskip}     \fi
\ifx \shownote     \undefined \def \shownote      #1{#1}          \fi
\ifx \showarticletitle \undefined \def \showarticletitle #1{#1}   \fi
\ifx \showURL      \undefined \def \showURL       {\relax}        \fi
\providecommand\bibfield[2]{#2}
\providecommand\bibinfo[2]{#2}
\providecommand\natexlab[1]{#1}
\providecommand\showeprint[2][]{arXiv:#2}

\bibitem[Alahi et~al\mbox{.}(2016)]%
        {alahi2016social}
\bibfield{author}{\bibinfo{person}{Alexandre Alahi}, \bibinfo{person}{Kratarth
  Goel}, \bibinfo{person}{Vignesh Ramanathan}, \bibinfo{person}{Alexandre
  Robicquet}, \bibinfo{person}{Li Fei-Fei}, {and} \bibinfo{person}{Silvio
  Savarese}.} \bibinfo{year}{2016}\natexlab{}.
\newblock \showarticletitle{Social lstm: Human trajectory prediction in crowded
  spaces}. In \bibinfo{booktitle}{\emph{Proceedings of the IEEE conference on
  computer vision and pattern recognition}}. \bibinfo{pages}{961--971}.
\newblock


\bibitem[An et~al\mbox{.}(2022)]%
        {dginet2022}
\bibfield{author}{\bibinfo{person}{Jiyao An}, \bibinfo{person}{Wei Liu},
  \bibinfo{person}{Qingqin Liu}, \bibinfo{person}{Liang Guo},
  \bibinfo{person}{Ping Ren}, {and} \bibinfo{person}{Tao Li}.}
  \bibinfo{year}{2022}\natexlab{}.
\newblock \showarticletitle{DGInet: Dynamic graph and interaction-aware
  convolutional network for vehicle trajectory prediction}.
\newblock \bibinfo{journal}{\emph{Neural Networks}}  \bibinfo{volume}{151}
  (\bibinfo{year}{2022}), \bibinfo{pages}{336--348}.
\newblock


\bibitem[Awad~George(2018)]%
        {awad2018benchmarking}
\bibfield{author}{\bibinfo{person}{TRECVID Awad~George}.}
  \bibinfo{year}{2018}\natexlab{}.
\newblock \showarticletitle{Benchmarking Video Activity Detection, Video
  Captioning and Matching, Video Storytelling Linking and Video Search}. In
  \bibinfo{booktitle}{\emph{Proceedings of TRECVID}}.
\newblock


\bibitem[Brkljač et~al\mbox{.}(2022)]%
        {9797078}
\bibfield{author}{\bibinfo{person}{Branko Brkljač}, \bibinfo{person}{Boris
  Antić}, {and} \bibinfo{person}{Zoran Mitrović}.}
  \bibinfo{year}{2022}\natexlab{}.
\newblock \showarticletitle{Potential of embedded vision platforms in
  development of spatial AI enabled CPS}. In \bibinfo{booktitle}{\emph{2022
  11th Mediterranean Conference on Embedded Computing (MECO)}}.
  \bibinfo{pages}{1--4}.
\newblock
\urldef\tempurl%
\url{https://doi.org/10.1109/MECO55406.2022.9797078}
\showDOI{\tempurl}


\bibitem[Cai et~al\mbox{.}(2020)]%
        {cai2020context}
\bibfield{author}{\bibinfo{person}{Jiannan Cai}, \bibinfo{person}{Yuxi Zhang},
  \bibinfo{person}{Liu Yang}, \bibinfo{person}{Hubo Cai}, {and}
  \bibinfo{person}{Shuai Li}.} \bibinfo{year}{2020}\natexlab{}.
\newblock \showarticletitle{A context-augmented deep learning approach for
  worker trajectory prediction on unstructured and dynamic construction sites}.
\newblock \bibinfo{journal}{\emph{Advanced Engineering Informatics}}
  \bibinfo{volume}{46} (\bibinfo{year}{2020}), \bibinfo{pages}{101173}.
\newblock


\bibitem[Chen et~al\mbox{.}(2020)]%
        {Chen2020WITM}
\bibfield{author}{\bibinfo{person}{Caijuan Chen}, \bibinfo{person}{Kaoru Ota},
  \bibinfo{person}{Mianxiong Dong}, \bibinfo{person}{Chen Yu}, {and}
  \bibinfo{person}{Hai Jin}.} \bibinfo{year}{2020}\natexlab{}.
\newblock \showarticletitle{WITM: Intelligent Traffic Monitoring Using
  Fine-Grained Wireless Signal}.
\newblock \bibinfo{journal}{\emph{IEEE Transactions on Emerging Topics in
  Computational Intelligence}} \bibinfo{volume}{4}, \bibinfo{number}{3}
  (\bibinfo{year}{2020}), \bibinfo{pages}{206--215}.
\newblock
\urldef\tempurl%
\url{https://doi.org/10.1109/TETCI.2019.2926505}
\showDOI{\tempurl}


\bibitem[Chen(2017)]%
        {chen2017research}
\bibfield{author}{\bibinfo{person}{Dawei Chen}.}
  \bibinfo{year}{2017}\natexlab{}.
\newblock \showarticletitle{Research on traffic flow prediction in the big data
  environment based on the improved RBF neural network}.
\newblock \bibinfo{journal}{\emph{IEEE Transactions on Industrial Informatics}}
  \bibinfo{volume}{13}, \bibinfo{number}{4} (\bibinfo{year}{2017}),
  \bibinfo{pages}{2000--2008}.
\newblock


\bibitem[Chen et~al\mbox{.}(2021)]%
        {Chen_2021_ICCV}
\bibfield{author}{\bibinfo{person}{Guangyi Chen}, \bibinfo{person}{Junlong Li},
  \bibinfo{person}{Jiwen Lu}, {and} \bibinfo{person}{Jie Zhou}.}
  \bibinfo{year}{2021}\natexlab{}.
\newblock \showarticletitle{Human Trajectory Prediction via Counterfactual
  Analysis}. In \bibinfo{booktitle}{\emph{Proceedings of the IEEE/CVF
  International Conference on Computer Vision (ICCV)}}.
  \bibinfo{pages}{9824--9833}.
\newblock


\bibitem[Chen and Demachi(2021)]%
        {Chen2021TowardsOH}
\bibfield{author}{\bibinfo{person}{Shi Chen} {and} \bibinfo{person}{Kazuyuki
  Demachi}.} \bibinfo{year}{2021}\natexlab{}.
\newblock \showarticletitle{Towards on-site hazards identification of improper
  use of personal protective equipment using deep learning-based geometric
  relationships and hierarchical scene graph}.
\newblock \bibinfo{journal}{\emph{Automation in Construction}}
  \bibinfo{volume}{125} (\bibinfo{year}{2021}), \bibinfo{pages}{103619}.
\newblock


\bibitem[Chen et~al\mbox{.}(2022)]%
        {chen2022scept}
\bibfield{author}{\bibinfo{person}{Yuxiao Chen}, \bibinfo{person}{Boris
  Ivanovic}, {and} \bibinfo{person}{Marco Pavone}.}
  \bibinfo{year}{2022}\natexlab{}.
\newblock \showarticletitle{ScePT: Scene-Consistent, Policy-Based Trajectory
  Predictions for Planning}. In \bibinfo{booktitle}{\emph{Proceedings of the
  IEEE/CVF Conference on Computer Vision and Pattern Recognition}}.
  \bibinfo{pages}{17103--17112}.
\newblock


\bibitem[Chwa et~al\mbox{.}(2018)]%
        {chwa2018closing}
\bibfield{author}{\bibinfo{person}{Hoon~Sung Chwa}, \bibinfo{person}{Kang~G
  Shin}, {and} \bibinfo{person}{Jinkyu Lee}.} \bibinfo{year}{2018}\natexlab{}.
\newblock \showarticletitle{Closing the gap between stability and
  schedulability: A new task model for cyber-physical systems}. In
  \bibinfo{booktitle}{\emph{2018 IEEE Real-Time and Embedded Technology and
  Applications Symposium (RTAS)}}. IEEE, \bibinfo{pages}{327--337}.
\newblock


\bibitem[Colyar and Halkias(2006)]%
        {NGSIM_i80}
\bibfield{author}{\bibinfo{person}{James Colyar} {and} \bibinfo{person}{John
  Halkias}.} \bibinfo{year}{2006}\natexlab{}.
\newblock \bibinfo{title}{Next Generation SIMulation ({NGSIM}), {I}nterstate 80
  Freeway Dataset. {FHWA-HRT}-06-137}.
\newblock
\newblock
\urldef\tempurl%
\url{https://www.fhwa.dot.gov/publications/research/operations/06137/}
\showURL{%
\tempurl}


\bibitem[Colyar and Halkias(2007)]%
        {NGSIM_US101}
\bibfield{author}{\bibinfo{person}{James Colyar} {and} \bibinfo{person}{John
  Halkias}.} \bibinfo{year}{2007}\natexlab{}.
\newblock \bibinfo{title}{Next Generation SIMulation ({NGSIM}), {US}
  {H}ighway-101 dataset. {FHWA-HRT}-07-030.}
\newblock
\newblock
\urldef\tempurl%
\url{https://www.fhwa.dot.gov/publications/research/operations/07030/}
\showURL{%
\tempurl}


\bibitem[Cui et~al\mbox{.}(2019)]%
        {cui2019multimodal}
\bibfield{author}{\bibinfo{person}{Henggang Cui}, \bibinfo{person}{Vladan
  Radosavljevic}, \bibinfo{person}{Fang-Chieh Chou}, \bibinfo{person}{Tsung-Han
  Lin}, \bibinfo{person}{Thi Nguyen}, \bibinfo{person}{Tzu-Kuo Huang},
  \bibinfo{person}{Jeff Schneider}, {and} \bibinfo{person}{Nemanja Djuric}.}
  \bibinfo{year}{2019}\natexlab{}.
\newblock \showarticletitle{Multimodal trajectory predictions for autonomous
  driving using deep convolutional networks}. In \bibinfo{booktitle}{\emph{2019
  International Conference on Robotics and Automation (ICRA)}}. IEEE,
  \bibinfo{pages}{2090--2096}.
\newblock


\bibitem[D{\'e}niz~Cerpa et~al\mbox{.}(2022)]%
        {deniz2022efficient}
\bibfield{author}{\bibinfo{person}{Jos{\'e}~Daniel D{\'e}niz~Cerpa},
  \bibinfo{person}{Juan Isern}, \bibinfo{person}{Jan Solanti},
  \bibinfo{person}{Pekka J{\"a}{\"a}skel{\"a}inen}, \bibinfo{person}{Petr
  Hn{\v{e}}tynka}, \bibinfo{person}{Lubom{\'\i}r Bulej},
  \bibinfo{person}{Eduardo Ros~Vidal}, \bibinfo{person}{Francisco
  Barranco~Exp{\'o}sito}, {et~al\mbox{.}}} \bibinfo{year}{2022}\natexlab{}.
\newblock \showarticletitle{Efficient reconfigurable CPS for monitoring the
  elderly at home via Deep Learning}.
\newblock  (\bibinfo{year}{2022}).
\newblock


\bibitem[Deo and Trivedi(2018)]%
        {CSlstm18}
\bibfield{author}{\bibinfo{person}{Nachiket Deo} {and}
  \bibinfo{person}{Mohan~M. Trivedi}.} \bibinfo{year}{2018}\natexlab{}.
\newblock \showarticletitle{Convolutional Social Pooling for Vehicle Trajectory
  Prediction}. In \bibinfo{booktitle}{\emph{2018 {IEEE} Conference on Computer
  Vision and Pattern Recognition Workshops, {CVPR} Workshops 2018, Salt Lake
  City, UT, USA, June 18-22, 2018}}. \bibinfo{publisher}{Computer Vision
  Foundation / {IEEE} Computer Society}, \bibinfo{pages}{1468--1476}.
\newblock
\urldef\tempurl%
\url{https://doi.org/10.1109/CVPRW.2018.00196}
\showDOI{\tempurl}


\bibitem[Fang et~al\mbox{.}(2018)]%
        {fang2018detecting}
\bibfield{author}{\bibinfo{person}{Qi Fang}, \bibinfo{person}{Heng Li},
  \bibinfo{person}{Xiaochun Luo}, \bibinfo{person}{Lieyun Ding},
  \bibinfo{person}{Hanbin Luo}, \bibinfo{person}{Timothy~M Rose}, {and}
  \bibinfo{person}{Wangpeng An}.} \bibinfo{year}{2018}\natexlab{}.
\newblock \showarticletitle{Detecting non-hardhat-use by a deep learning method
  from far-field surveillance videos}.
\newblock \bibinfo{journal}{\emph{Automation in Construction}}
  \bibinfo{volume}{85} (\bibinfo{year}{2018}), \bibinfo{pages}{1--9}.
\newblock


\bibitem[Farooq et~al\mbox{.}(2021)]%
        {farooq2021flow}
\bibfield{author}{\bibinfo{person}{Basit Farooq}, \bibinfo{person}{Jinsong
  Bao}, \bibinfo{person}{Hanan Raza}, \bibinfo{person}{Yicheng Sun}, {and}
  \bibinfo{person}{Qingwen Ma}.} \bibinfo{year}{2021}\natexlab{}.
\newblock \showarticletitle{Flow-shop path planning for multi-automated guided
  vehicles in intelligent textile spinning cyber-physical production systems
  dynamic environment}.
\newblock \bibinfo{journal}{\emph{Journal of Manufacturing Systems}}
  \bibinfo{volume}{59} (\bibinfo{year}{2021}), \bibinfo{pages}{98--116}.
\newblock


\bibitem[Fu et~al\mbox{.}(2022)]%
        {highway2022}
\bibfield{author}{\bibinfo{person}{Mengyin Fu}, \bibinfo{person}{Ting Zhang},
  \bibinfo{person}{Wenjie Song}, \bibinfo{person}{Yi Yang}, {and}
  \bibinfo{person}{Meiling Wang}.} \bibinfo{year}{2022}\natexlab{}.
\newblock \showarticletitle{Trajectory Prediction-Based Local Spatio-Temporal
  Navigation Map for Autonomous Driving in Dynamic Highway Environments}.
\newblock \bibinfo{journal}{\emph{{IEEE} Trans. Intell. Transp. Syst.}}
  \bibinfo{volume}{23}, \bibinfo{number}{7} (\bibinfo{year}{2022}),
  \bibinfo{pages}{6418--6429}.
\newblock
\urldef\tempurl%
\url{https://doi.org/10.1109/TITS.2021.3057110}
\showDOI{\tempurl}


\bibitem[Georgiou et~al\mbox{.}(2018)]%
        {georgiou2018moving}
\bibfield{author}{\bibinfo{person}{Harris Georgiou}, \bibinfo{person}{Sophia
  Karagiorgou}, \bibinfo{person}{Yannis Kontoulis}, \bibinfo{person}{Nikos
  Pelekis}, \bibinfo{person}{Petros Petrou}, \bibinfo{person}{David Scarlatti},
  {and} \bibinfo{person}{Yannis Theodoridis}.} \bibinfo{year}{2018}\natexlab{}.
\newblock \showarticletitle{Moving objects analytics: Survey on future location
  \& trajectory prediction methods}.
\newblock \bibinfo{journal}{\emph{arXiv preprint arXiv:1807.04639}}
  (\bibinfo{year}{2018}).
\newblock


\bibitem[Gupta et~al\mbox{.}(2018)]%
        {gupta2018social}
\bibfield{author}{\bibinfo{person}{Agrim Gupta}, \bibinfo{person}{Justin
  Johnson}, \bibinfo{person}{Li Fei-Fei}, \bibinfo{person}{Silvio Savarese},
  {and} \bibinfo{person}{Alexandre Alahi}.} \bibinfo{year}{2018}\natexlab{}.
\newblock \showarticletitle{Social gan: Socially acceptable trajectories with
  generative adversarial networks}. In \bibinfo{booktitle}{\emph{Proceedings of
  the IEEE conference on computer vision and pattern recognition}}.
  \bibinfo{pages}{2255--2264}.
\newblock


\bibitem[Hussain et~al\mbox{.}(2022)]%
        {hussain2022vision}
\bibfield{author}{\bibinfo{person}{Manzoor Hussain}, \bibinfo{person}{Nazakat
  Ali}, {and} \bibinfo{person}{Jang-Eui Hong}.}
  \bibinfo{year}{2022}\natexlab{}.
\newblock \showarticletitle{Vision beyond the field-of-view: a collaborative
  perception system to improve safety of intelligent cyber-physical systems}.
\newblock \bibinfo{journal}{\emph{Sensors}} \bibinfo{volume}{22},
  \bibinfo{number}{17} (\bibinfo{year}{2022}), \bibinfo{pages}{6610}.
\newblock


\bibitem[Isern et~al\mbox{.}(2020)]%
        {Isern2020Reconfig}
\bibfield{author}{\bibinfo{person}{Juan Isern}, \bibinfo{person}{Francisco
  Barranco}, \bibinfo{person}{Daniel Deniz}, \bibinfo{person}{Juho Lesonen},
  \bibinfo{person}{Jari Hannuksela}, {and} \bibinfo{person}{Richard~R.
  Carrillo}.} \bibinfo{year}{2020}\natexlab{}.
\newblock \showarticletitle{Reconfigurable cyber-physical system for critical
  infrastructure protection in smart cities via smart video-surveillance}.
\newblock \bibinfo{journal}{\emph{Pattern Recognit. Lett.}}
  \bibinfo{volume}{140} (\bibinfo{year}{2020}), \bibinfo{pages}{303--309}.
\newblock
\urldef\tempurl%
\url{https://doi.org/10.1016/j.patrec.2020.11.004}
\showDOI{\tempurl}


\bibitem[Jeon et~al\mbox{.}(2020)]%
        {jeon2020scale}
\bibfield{author}{\bibinfo{person}{Hyeongseok Jeon}, \bibinfo{person}{Junwon
  Choi}, {and} \bibinfo{person}{Dongsuk Kum}.} \bibinfo{year}{2020}\natexlab{}.
\newblock \showarticletitle{Scale-net: Scalable vehicle trajectory prediction
  network under random number of interacting vehicles via edge-enhanced graph
  convolutional neural network}. In \bibinfo{booktitle}{\emph{2020 IEEE/RSJ
  International Conference on Intelligent Robots and Systems (IROS)}}. IEEE,
  \bibinfo{pages}{2095--2102}.
\newblock


\bibitem[Jeong et~al\mbox{.}(2013)]%
        {Jeong2013traffFlow}
\bibfield{author}{\bibinfo{person}{Young-Seon Jeong}, \bibinfo{person}{Young-Ji
  Byon}, \bibinfo{person}{Manoel~Mendonca Castro-Neto}, {and}
  \bibinfo{person}{Said~M. Easa}.} \bibinfo{year}{2013}\natexlab{}.
\newblock \showarticletitle{Supervised Weighting-Online Learning Algorithm for
  Short-Term Traffic Flow Prediction}.
\newblock \bibinfo{journal}{\emph{IEEE Transactions on Intelligent
  Transportation Systems}} \bibinfo{volume}{14}, \bibinfo{number}{4}
  (\bibinfo{year}{2013}), \bibinfo{pages}{1700--1707}.
\newblock
\urldef\tempurl%
\url{https://doi.org/10.1109/TITS.2013.2267735}
\showDOI{\tempurl}


\bibitem[Jones et~al\mbox{.}(2014)]%
        {jones2014anomaly}
\bibfield{author}{\bibinfo{person}{Austin Jones}, \bibinfo{person}{Zhaodan
  Kong}, {and} \bibinfo{person}{Calin Belta}.} \bibinfo{year}{2014}\natexlab{}.
\newblock \showarticletitle{Anomaly detection in cyber-physical systems: A
  formal methods approach}. In \bibinfo{booktitle}{\emph{53rd IEEE Conference
  on Decision and Control}}. \bibinfo{pages}{848--853}.
\newblock
\urldef\tempurl%
\url{https://doi.org/10.1109/CDC.2014.7039487}
\showDOI{\tempurl}


\bibitem[Katariya et~al\mbox{.}(2022)]%
        {katariya2021deeptrack}
\bibfield{author}{\bibinfo{person}{Vinit Katariya},
  \bibinfo{person}{Mohammadreza Baharani}, \bibinfo{person}{Nichole Morris},
  \bibinfo{person}{Omidreza Shoghli}, {and} \bibinfo{person}{Hamed Tabkhi}.}
  \bibinfo{year}{2022}\natexlab{}.
\newblock \showarticletitle{DeepTrack: Lightweight Deep Learning for Vehicle
  Trajectory Prediction in Highways}.
\newblock \bibinfo{journal}{\emph{IEEE Transactions on Intelligent
  Transportation Systems}} (\bibinfo{year}{2022}), \bibinfo{pages}{1--10}.
\newblock
\urldef\tempurl%
\url{https://doi.org/10.1109/TITS.2022.3172015}
\showDOI{\tempurl}


\bibitem[Kaur and Chatterjee(2022)]%
        {kaur2022applications}
\bibfield{author}{\bibinfo{person}{Amandeep Kaur} {and}
  \bibinfo{person}{Jyotir~Moy Chatterjee}.} \bibinfo{year}{2022}\natexlab{}.
\newblock \showarticletitle{Applications of Cyber-Physical Systems}.
\newblock \bibinfo{journal}{\emph{Cyber-Physical Systems: Foundations and
  Techniques}} (\bibinfo{year}{2022}), \bibinfo{pages}{289--310}.
\newblock


\bibitem[Kingma and Ba(2014)]%
        {kingma2014adam}
\bibfield{author}{\bibinfo{person}{Diederik~P Kingma} {and}
  \bibinfo{person}{Jimmy Ba}.} \bibinfo{year}{2014}\natexlab{}.
\newblock \showarticletitle{Adam: A method for stochastic optimization}.
\newblock \bibinfo{journal}{\emph{arXiv preprint arXiv:1412.6980}}
  (\bibinfo{year}{2014}).
\newblock


\bibitem[Kipf and Welling(2016)]%
        {kipf2016semi}
\bibfield{author}{\bibinfo{person}{Thomas~N Kipf} {and} \bibinfo{person}{Max
  Welling}.} \bibinfo{year}{2016}\natexlab{}.
\newblock \showarticletitle{Semi-supervised classification with graph
  convolutional networks}.
\newblock \bibinfo{journal}{\emph{arXiv preprint arXiv:1609.02907}}
  (\bibinfo{year}{2016}).
\newblock


\bibitem[Leman and Weisfeiler(1968)]%
        {leman1968reduction}
\bibfield{author}{\bibinfo{person}{AA Leman} {and} \bibinfo{person}{Boris
  Weisfeiler}.} \bibinfo{year}{1968}\natexlab{}.
\newblock \showarticletitle{A reduction of a graph to a canonical form and an
  algebra arising during this reduction}.
\newblock \bibinfo{journal}{\emph{Nauchno-Technicheskaya Informatsiya}}
  \bibinfo{volume}{2}, \bibinfo{number}{9} (\bibinfo{year}{1968}),
  \bibinfo{pages}{12--16}.
\newblock


\bibitem[Lerner et~al\mbox{.}(2007)]%
        {lerner2007crowds}
\bibfield{author}{\bibinfo{person}{Alon Lerner}, \bibinfo{person}{Yiorgos
  Chrysanthou}, {and} \bibinfo{person}{Dani Lischinski}.}
  \bibinfo{year}{2007}\natexlab{}.
\newblock \showarticletitle{Crowds by example}. In
  \bibinfo{booktitle}{\emph{Computer graphics forum}},
  Vol.~\bibinfo{volume}{26}. Wiley Online Library, \bibinfo{pages}{655--664}.
\newblock


\bibitem[Li et~al\mbox{.}(2022)]%
        {li2022graph}
\bibfield{author}{\bibinfo{person}{Lihuan Li}, \bibinfo{person}{Maurice
  Pagnucco}, {and} \bibinfo{person}{Yang Song}.}
  \bibinfo{year}{2022}\natexlab{}.
\newblock \showarticletitle{Graph-Based Spatial Transformer With Memory Replay
  for Multi-Future Pedestrian Trajectory Prediction}. In
  \bibinfo{booktitle}{\emph{Proceedings of the IEEE/CVF Conference on Computer
  Vision and Pattern Recognition}}. \bibinfo{pages}{2231--2241}.
\newblock


\bibitem[Li et~al\mbox{.}(2019)]%
        {grip2019}
\bibfield{author}{\bibinfo{person}{Xin Li}, \bibinfo{person}{Xiaowen Ying},
  {and} \bibinfo{person}{Mooi~Choo Chuah}.} \bibinfo{year}{2019}\natexlab{}.
\newblock \showarticletitle{{GRIP:} Graph-based Interaction-aware Trajectory
  Prediction}. In \bibinfo{booktitle}{\emph{2019 {IEEE} Intelligent
  Transportation Systems Conference, {ITSC} 2019, Auckland, New Zealand,
  October 27-30, 2019}}. \bibinfo{publisher}{{IEEE}},
  \bibinfo{pages}{3960--3966}.
\newblock
\urldef\tempurl%
\url{https://doi.org/10.1109/ITSC.2019.8917228}
\showDOI{\tempurl}


\bibitem[Li et~al\mbox{.}(2020)]%
        {2020gripplus}
\bibfield{author}{\bibinfo{person}{Xin Li}, \bibinfo{person}{Xiaowen Ying},
  {and} \bibinfo{person}{Mooi~Choo Chuah}.} \bibinfo{year}{2020}\natexlab{}.
\newblock \showarticletitle{GRIP++: Enhanced Graph-based Interaction-aware
  Trajectory Prediction for Autonomous Driving}.
\newblock \bibinfo{journal}{\emph{arXiv preprint arXiv:1907.07792}}
  (\bibinfo{year}{2020}).
\newblock


\bibitem[Lian et~al\mbox{.}(2020)]%
        {lian2020cyber}
\bibfield{author}{\bibinfo{person}{Yindong Lian}, \bibinfo{person}{Qifan Yang},
  \bibinfo{person}{Wei Xie}, {and} \bibinfo{person}{Langwen Zhang}.}
  \bibinfo{year}{2020}\natexlab{}.
\newblock \showarticletitle{Cyber-physical system-based heuristic planning and
  scheduling method for multiple automatic guided vehicles in logistics
  systems}.
\newblock \bibinfo{journal}{\emph{IEEE Transactions on Industrial Informatics}}
  \bibinfo{volume}{17}, \bibinfo{number}{11} (\bibinfo{year}{2020}),
  \bibinfo{pages}{7882--7893}.
\newblock


\bibitem[Liang et~al\mbox{.}(2020a)]%
        {liang2020simaug}
\bibfield{author}{\bibinfo{person}{Junwei Liang}, \bibinfo{person}{Lu Jiang},
  {and} \bibinfo{person}{Alexander Hauptmann}.}
  \bibinfo{year}{2020}\natexlab{a}.
\newblock \showarticletitle{Simaug: Learning robust representations from
  simulation for trajectory prediction}. In \bibinfo{booktitle}{\emph{European
  Conference on Computer Vision}}. Springer, \bibinfo{pages}{275--292}.
\newblock


\bibitem[Liang et~al\mbox{.}(2020b)]%
        {liang2020garden}
\bibfield{author}{\bibinfo{person}{Junwei Liang}, \bibinfo{person}{Lu Jiang},
  \bibinfo{person}{Kevin Murphy}, \bibinfo{person}{Ting Yu}, {and}
  \bibinfo{person}{Alexander Hauptmann}.} \bibinfo{year}{2020}\natexlab{b}.
\newblock \showarticletitle{The garden of forking paths: Towards multi-future
  trajectory prediction}. In \bibinfo{booktitle}{\emph{Proceedings of the
  IEEE/CVF Conference on Computer Vision and Pattern Recognition}}.
  \bibinfo{pages}{10508--10518}.
\newblock


\bibitem[Liang et~al\mbox{.}(2019)]%
        {liang2019peeking}
\bibfield{author}{\bibinfo{person}{Junwei Liang}, \bibinfo{person}{Lu Jiang},
  \bibinfo{person}{Juan~Carlos Niebles}, \bibinfo{person}{Alexander~G
  Hauptmann}, {and} \bibinfo{person}{Li Fei-Fei}.}
  \bibinfo{year}{2019}\natexlab{}.
\newblock \showarticletitle{Peeking into the future: Predicting future person
  activities and locations in videos}. In \bibinfo{booktitle}{\emph{Proceedings
  of the IEEE/CVF conference on computer vision and pattern recognition}}.
  \bibinfo{pages}{5725--5734}.
\newblock


\bibitem[Lin et~al\mbox{.}(2022)]%
        {STALSTM2021}
\bibfield{author}{\bibinfo{person}{Lei Lin}, \bibinfo{person}{Weizi Li},
  \bibinfo{person}{Huikun Bi}, {and} \bibinfo{person}{Lingqiao Qin}.}
  \bibinfo{year}{2022}\natexlab{}.
\newblock \showarticletitle{Vehicle Trajectory Prediction Using LSTMs With
  Spatial-Temporal Attention Mechanisms}.
\newblock \bibinfo{journal}{\emph{{IEEE} Intell. Transp. Syst. Mag.}}
  \bibinfo{volume}{14}, \bibinfo{number}{2} (\bibinfo{year}{2022}),
  \bibinfo{pages}{197--208}.
\newblock
\urldef\tempurl%
\url{https://doi.org/10.1109/MITS.2021.3049404}
\showDOI{\tempurl}


\bibitem[Liu et~al\mbox{.}(2022)]%
        {liu2022multi}
\bibfield{author}{\bibinfo{person}{Yongkang Liu}, \bibinfo{person}{Xuewei Qi},
  \bibinfo{person}{Emrah~Akin Sisbot}, {and} \bibinfo{person}{Kentaro Oguchi}.}
  \bibinfo{year}{2022}\natexlab{}.
\newblock \showarticletitle{Multi-Agent Trajectory Prediction with Graph
  Attention Isomorphism Neural Network}. In \bibinfo{booktitle}{\emph{2022 IEEE
  Intelligent Vehicles Symposium (IV)}}. IEEE, \bibinfo{pages}{273--279}.
\newblock


\bibitem[Lv et~al\mbox{.}(2021)]%
        {lv2021ssagcn}
\bibfield{author}{\bibinfo{person}{Pei Lv}, \bibinfo{person}{Wentong Wang},
  \bibinfo{person}{Yunxin Wang}, \bibinfo{person}{Yuzhen Zhang},
  \bibinfo{person}{Mingliang Xu}, {and} \bibinfo{person}{Changsheng Xu}.}
  \bibinfo{year}{2021}\natexlab{}.
\newblock \showarticletitle{SSAGCN: Social Soft Attention Graph Convolution
  Network for Pedestrian Trajectory Prediction}.
\newblock \bibinfo{journal}{\emph{arXiv preprint arXiv:2112.02459}}
  (\bibinfo{year}{2021}).
\newblock


\bibitem[Mangalam et~al\mbox{.}(2021)]%
        {mangalam2021goals}
\bibfield{author}{\bibinfo{person}{Karttikeya Mangalam}, \bibinfo{person}{Yang
  An}, \bibinfo{person}{Harshayu Girase}, {and} \bibinfo{person}{Jitendra
  Malik}.} \bibinfo{year}{2021}\natexlab{}.
\newblock \showarticletitle{From goals, waypoints \& paths to long term human
  trajectory forecasting}. In \bibinfo{booktitle}{\emph{Proceedings of the
  IEEE/CVF International Conference on Computer Vision}}.
  \bibinfo{pages}{15233--15242}.
\newblock


\bibitem[Mendieta and Tabkhi(2021)]%
        {mendieta2021carpe}
\bibfield{author}{\bibinfo{person}{Mat{\'\i}as Mendieta} {and}
  \bibinfo{person}{Hamed Tabkhi}.} \bibinfo{year}{2021}\natexlab{}.
\newblock \showarticletitle{CARPe posterum: A convolutional approach for
  real-time pedestrian path prediction}. In
  \bibinfo{booktitle}{\emph{Proceedings of the AAAI Conference on Artificial
  Intelligence}}, Vol.~\bibinfo{volume}{35}. \bibinfo{pages}{2346--2354}.
\newblock


\bibitem[Mo et~al\mbox{.}(2022)]%
        {HEAT}
\bibfield{author}{\bibinfo{person}{Xiaoyu Mo}, \bibinfo{person}{Zhiyu Huang},
  \bibinfo{person}{Yang Xing}, {and} \bibinfo{person}{Chen Lv}.}
  \bibinfo{year}{2022}\natexlab{}.
\newblock \showarticletitle{Multi-Agent Trajectory Prediction With
  Heterogeneous Edge-Enhanced Graph Attention Network}.
\newblock \bibinfo{journal}{\emph{IEEE Transactions on Intelligent
  Transportation Systems}} \bibinfo{volume}{23}, \bibinfo{number}{7}
  (\bibinfo{year}{2022}), \bibinfo{pages}{9554--9567}.
\newblock
\urldef\tempurl%
\url{https://doi.org/10.1109/TITS.2022.3146300}
\showDOI{\tempurl}


\bibitem[Niu et~al\mbox{.}(2021)]%
        {NIU202148}
\bibfield{author}{\bibinfo{person}{Zhaoyang Niu}, \bibinfo{person}{Guoqiang
  Zhong}, {and} \bibinfo{person}{Hui Yu}.} \bibinfo{year}{2021}\natexlab{}.
\newblock \showarticletitle{A review on the attention mechanism of deep
  learning}.
\newblock \bibinfo{journal}{\emph{Neurocomputing}}  \bibinfo{volume}{452}
  (\bibinfo{year}{2021}), \bibinfo{pages}{48--62}.
\newblock
\showISSN{0925-2312}
\urldef\tempurl%
\url{https://doi.org/10.1016/j.neucom.2021.03.091}
\showDOI{\tempurl}


\bibitem[Pasandideh et~al\mbox{.}(2022)]%
        {9757168}
\bibfield{author}{\bibinfo{person}{Shabnam Pasandideh}, \bibinfo{person}{Pedro
  Pereira}, {and} \bibinfo{person}{Luis Gomes}.}
  \bibinfo{year}{2022}\natexlab{}.
\newblock \showarticletitle{Cyber-Physical-Social Systems: Taxonomy,
  Challenges, and Opportunities}.
\newblock \bibinfo{journal}{\emph{IEEE Access}}  \bibinfo{volume}{10}
  (\bibinfo{year}{2022}), \bibinfo{pages}{42404--42419}.
\newblock
\urldef\tempurl%
\url{https://doi.org/10.1109/ACCESS.2022.3167441}
\showDOI{\tempurl}


\bibitem[Pazho et~al\mbox{.}(2023)]%
        {pazho2023ancilia}
\bibfield{author}{\bibinfo{person}{Armin~Danesh Pazho},
  \bibinfo{person}{Christopher Neff}, \bibinfo{person}{Ghazal~Alinezhad
  Noghre}, \bibinfo{person}{Babak~Rahimi Ardabili}, \bibinfo{person}{Shanle
  Yao}, \bibinfo{person}{Mohammadreza Baharani}, {and} \bibinfo{person}{Hamed
  Tabkhi}.} \bibinfo{year}{2023}\natexlab{}.
\newblock \showarticletitle{Ancilia: Scalable Intelligent Video Surveillance
  for the Artificial Intelligence of Things}.
\newblock \bibinfo{journal}{\emph{arXiv preprint arXiv:2301.03561}}
  (\bibinfo{year}{2023}).
\newblock


\bibitem[Pellegrini et~al\mbox{.}(2009)]%
        {pellegrini2009you}
\bibfield{author}{\bibinfo{person}{Stefano Pellegrini},
  \bibinfo{person}{Andreas Ess}, \bibinfo{person}{Konrad Schindler}, {and}
  \bibinfo{person}{Luc Van~Gool}.} \bibinfo{year}{2009}\natexlab{}.
\newblock \showarticletitle{You'll never walk alone: Modeling social behavior
  for multi-target tracking}. In \bibinfo{booktitle}{\emph{2009 IEEE 12th
  international conference on computer vision}}. IEEE,
  \bibinfo{pages}{261--268}.
\newblock


\bibitem[Peng et~al\mbox{.}(2021)]%
        {peng2021sra}
\bibfield{author}{\bibinfo{person}{Yusheng Peng}, \bibinfo{person}{Gaofeng
  Zhang}, \bibinfo{person}{Jun Shi}, \bibinfo{person}{Benzhu Xu}, {and}
  \bibinfo{person}{Liping Zheng}.} \bibinfo{year}{2021}\natexlab{}.
\newblock \showarticletitle{SRA-LSTM: Social relationship attention LSTM for
  human trajectory prediction}.
\newblock \bibinfo{journal}{\emph{arXiv preprint arXiv:2103.17045}}
  (\bibinfo{year}{2021}).
\newblock


\bibitem[Pustokhina et~al\mbox{.}(2021)]%
        {pustokhina2021automated}
\bibfield{author}{\bibinfo{person}{Irina~V Pustokhina},
  \bibinfo{person}{Denis~A Pustokhin}, \bibinfo{person}{Thavavel Vaiyapuri},
  \bibinfo{person}{Deepak Gupta}, \bibinfo{person}{Sachin Kumar}, {and}
  \bibinfo{person}{K Shankar}.} \bibinfo{year}{2021}\natexlab{}.
\newblock \showarticletitle{An automated deep learning based anomaly detection
  in pedestrian walkways for vulnerable road users safety}.
\newblock \bibinfo{journal}{\emph{Safety science}}  \bibinfo{volume}{142}
  (\bibinfo{year}{2021}), \bibinfo{pages}{105356}.
\newblock


\bibitem[Sabeti et~al\mbox{.}(2021)]%
        {sabeti2021toward}
\bibfield{author}{\bibinfo{person}{Sepehr Sabeti}, \bibinfo{person}{Omidreza
  Shoghli}, \bibinfo{person}{Mohammadreza Baharani}, {and}
  \bibinfo{person}{Hamed Tabkhi}.} \bibinfo{year}{2021}\natexlab{}.
\newblock \showarticletitle{Toward AI-enabled augmented reality to enhance the
  safety of highway work zones: Feasibility, requirements, and challenges}.
\newblock \bibinfo{journal}{\emph{Advanced Engineering Informatics}}
  \bibinfo{volume}{50} (\bibinfo{year}{2021}), \bibinfo{pages}{101429}.
\newblock


\bibitem[Salzmann et~al\mbox{.}(2020)]%
        {salzmann2020trajectron++}
\bibfield{author}{\bibinfo{person}{Tim Salzmann}, \bibinfo{person}{Boris
  Ivanovic}, \bibinfo{person}{Punarjay Chakravarty}, {and}
  \bibinfo{person}{Marco Pavone}.} \bibinfo{year}{2020}\natexlab{}.
\newblock \showarticletitle{Trajectron++: Dynamically-feasible trajectory
  forecasting with heterogeneous data}. In \bibinfo{booktitle}{\emph{European
  Conference on Computer Vision}}. Springer, \bibinfo{pages}{683--700}.
\newblock


\bibitem[Sanchez et~al\mbox{.}(2021)]%
        {sanchez2021real}
\bibfield{author}{\bibinfo{person}{Justin Sanchez},
  \bibinfo{person}{Christopher Neff}, {and} \bibinfo{person}{Hamed Tabkhi}.}
  \bibinfo{year}{2021}\natexlab{}.
\newblock \showarticletitle{Real-world graph convolution networks (RW-GCNS) for
  action recognition in smart video surveillance}. In
  \bibinfo{booktitle}{\emph{2021 IEEE/ACM Symposium on Edge Computing (SEC)}}.
  IEEE, \bibinfo{pages}{121--134}.
\newblock


\bibitem[Shafiee et~al\mbox{.}(2021)]%
        {Shafiee_2021_CVPR}
\bibfield{author}{\bibinfo{person}{Nasim Shafiee}, \bibinfo{person}{Taskin
  Padir}, {and} \bibinfo{person}{Ehsan Elhamifar}.}
  \bibinfo{year}{2021}\natexlab{}.
\newblock \showarticletitle{Introvert: Human Trajectory Prediction via
  Conditional 3D Attention}. In \bibinfo{booktitle}{\emph{Proceedings of the
  IEEE/CVF Conference on Computer Vision and Pattern Recognition (CVPR)}}.
  \bibinfo{pages}{16815--16825}.
\newblock


\bibitem[Sheng et~al\mbox{.}(2021)]%
        {GSTCN2022}
\bibfield{author}{\bibinfo{person}{Zihao Sheng}, \bibinfo{person}{Yunwen Xu},
  \bibinfo{person}{Shibei Xue}, {and} \bibinfo{person}{Dewei Li}.}
  \bibinfo{year}{2021}\natexlab{}.
\newblock \showarticletitle{Graph-Based Spatial-Temporal Convolutional Network
  for Vehicle Trajectory Prediction in Autonomous Driving}.
\newblock \bibinfo{journal}{\emph{CoRR}}  \bibinfo{volume}{abs/2109.12764}
  (\bibinfo{year}{2021}).
\newblock
\showeprint[arXiv]{2109.12764}
\urldef\tempurl%
\url{https://arxiv.org/abs/2109.12764}
\showURL{%
\tempurl}


\bibitem[Song et~al\mbox{.}(2020)]%
        {Pip2021}
\bibfield{author}{\bibinfo{person}{Haoran Song}, \bibinfo{person}{Wenchao
  Ding}, \bibinfo{person}{Yuxuan Chen}, \bibinfo{person}{Shaojie Shen},
  \bibinfo{person}{Michael~Yu Wang}, {and} \bibinfo{person}{Qifeng Chen}.}
  \bibinfo{year}{2020}\natexlab{}.
\newblock \showarticletitle{PiP: Planning-Informed Trajectory Prediction for
  Autonomous Driving}. In \bibinfo{booktitle}{\emph{Computer Vision - {ECCV}
  2020 - 16th European Conference, Glasgow, UK, August 23-28, 2020,
  Proceedings, Part {XXI}}} \emph{(\bibinfo{series}{Lecture Notes in Computer
  Science}, Vol.~\bibinfo{volume}{12366})},
  \bibfield{editor}{\bibinfo{person}{Andrea Vedaldi}, \bibinfo{person}{Horst
  Bischof}, \bibinfo{person}{Thomas Brox}, {and} \bibinfo{person}{Jan{-}Michael
  Frahm}} (Eds.). \bibinfo{publisher}{Springer}, \bibinfo{pages}{598--614}.
\newblock
\urldef\tempurl%
\url{https://doi.org/10.1007/978-3-030-58589-1\_36}
\showDOI{\tempurl}


\bibitem[Veli{\v{c}}kovi{\'c} et~al\mbox{.}(2017)]%
        {velivckovic2017graph}
\bibfield{author}{\bibinfo{person}{Petar Veli{\v{c}}kovi{\'c}},
  \bibinfo{person}{Guillem Cucurull}, \bibinfo{person}{Arantxa Casanova},
  \bibinfo{person}{Adriana Romero}, \bibinfo{person}{Pietro Lio}, {and}
  \bibinfo{person}{Yoshua Bengio}.} \bibinfo{year}{2017}\natexlab{}.
\newblock \showarticletitle{Graph attention networks}.
\newblock \bibinfo{journal}{\emph{arXiv preprint arXiv:1710.10903}}
  (\bibinfo{year}{2017}).
\newblock


\bibitem[Vemula et~al\mbox{.}(2017)]%
        {vemula2017social}
\bibfield{author}{\bibinfo{person}{Anirudh Vemula}, \bibinfo{person}{Katharina
  Muelling}, {and} \bibinfo{person}{Jean Oh}.} \bibinfo{year}{2017}\natexlab{}.
\newblock \bibinfo{title}{Social Attention: Modeling Attention in Human
  Crowds}.
\newblock
\newblock
\urldef\tempurl%
\url{https://doi.org/10.48550/ARXIV.1710.04689}
\showDOI{\tempurl}


\bibitem[Wang et~al\mbox{.}(2021)]%
        {Wang_2021_WACV}
\bibfield{author}{\bibinfo{person}{Chengxin Wang}, \bibinfo{person}{Shaofeng
  Cai}, {and} \bibinfo{person}{Gary Tan}.} \bibinfo{year}{2021}\natexlab{}.
\newblock \showarticletitle{GraphTCN: Spatio-Temporal Interaction Modeling for
  Human Trajectory Prediction}. In \bibinfo{booktitle}{\emph{Proceedings of the
  IEEE/CVF Winter Conference on Applications of Computer Vision (WACV)}}.
  \bibinfo{pages}{3450--3459}.
\newblock


\bibitem[Wang et~al\mbox{.}(2019)]%
        {wang2019exploring}
\bibfield{author}{\bibinfo{person}{Chujie Wang}, \bibinfo{person}{Lin Ma},
  \bibinfo{person}{Rongpeng Li}, \bibinfo{person}{Tariq~S Durrani}, {and}
  \bibinfo{person}{Honggang Zhang}.} \bibinfo{year}{2019}\natexlab{}.
\newblock \showarticletitle{Exploring trajectory prediction through machine
  learning methods}.
\newblock \bibinfo{journal}{\emph{IEEE Access}}  \bibinfo{volume}{7}
  (\bibinfo{year}{2019}), \bibinfo{pages}{101441--101452}.
\newblock


\bibitem[Wang et~al\mbox{.}(2022)]%
        {wang2022intelligent}
\bibfield{author}{\bibinfo{person}{Shuai Wang}, \bibinfo{person}{Xiaoyu Li},
  \bibinfo{person}{Wei Chen}, \bibinfo{person}{Weiqiang Fan}, {and}
  \bibinfo{person}{Zijian Tian}.} \bibinfo{year}{2022}\natexlab{}.
\newblock \showarticletitle{An Intelligent Vision-Based Method of Worker
  Identification for Industrial Internet of Things (IoT)}.
\newblock \bibinfo{journal}{\emph{Wireless Communications and Mobile
  Computing}}  \bibinfo{volume}{2022} (\bibinfo{year}{2022}).
\newblock


\bibitem[Wong et~al\mbox{.}(2021)]%
        {wong2021view}
\bibfield{author}{\bibinfo{person}{Conghao Wong}, \bibinfo{person}{Beihao Xia},
  \bibinfo{person}{Ziming Hong}, \bibinfo{person}{Qinmu Peng}, {and}
  \bibinfo{person}{Xinge You}.} \bibinfo{year}{2021}\natexlab{}.
\newblock \showarticletitle{View Vertically: A hierarchical network for
  trajectory prediction via fourier spectrums}.
\newblock \bibinfo{journal}{\emph{arXiv preprint arXiv:2110.07288}}
  (\bibinfo{year}{2021}).
\newblock


\bibitem[Woo et~al\mbox{.}(2018a)]%
        {cbam18}
\bibfield{author}{\bibinfo{person}{Sanghyun Woo}, \bibinfo{person}{Jongchan
  Park}, \bibinfo{person}{Joon{-}Young Lee}, {and} \bibinfo{person}{In~So
  Kweon}.} \bibinfo{year}{2018}\natexlab{a}.
\newblock \showarticletitle{{CBAM:} Convolutional Block Attention Module}. In
  \bibinfo{booktitle}{\emph{Computer Vision - {ECCV} 2018 - 15th European
  Conference, Munich, Germany, September 8-14, 2018, Proceedings, Part {VII}}}
  \emph{(\bibinfo{series}{Lecture Notes in Computer Science},
  Vol.~\bibinfo{volume}{11211})}, \bibfield{editor}{\bibinfo{person}{Vittorio
  Ferrari}, \bibinfo{person}{Martial Hebert}, \bibinfo{person}{Cristian
  Sminchisescu}, {and} \bibinfo{person}{Yair Weiss}} (Eds.).
  \bibinfo{publisher}{Springer}, \bibinfo{pages}{3--19}.
\newblock
\urldef\tempurl%
\url{https://doi.org/10.1007/978-3-030-01234-2\_1}
\showDOI{\tempurl}


\bibitem[Woo et~al\mbox{.}(2018b)]%
        {woo2018cbam}
\bibfield{author}{\bibinfo{person}{Sanghyun Woo}, \bibinfo{person}{Jongchan
  Park}, \bibinfo{person}{Joon-Young Lee}, {and} \bibinfo{person}{In~So
  Kweon}.} \bibinfo{year}{2018}\natexlab{b}.
\newblock \showarticletitle{Cbam: Convolutional block attention module}. In
  \bibinfo{booktitle}{\emph{Proceedings of the European conference on computer
  vision (ECCV)}}. \bibinfo{pages}{3--19}.
\newblock


\bibitem[Wu et~al\mbox{.}(2019)]%
        {wu2019deep}
\bibfield{author}{\bibinfo{person}{Lin Wu}, \bibinfo{person}{Brian~C Lovell},
  {and} \bibinfo{person}{Yang Wang}.} \bibinfo{year}{2019}\natexlab{}.
\newblock \showarticletitle{Deep Learning in Person Re-identification for
  Cyber-Physical Surveillance Systems}.
\newblock In \bibinfo{booktitle}{\emph{Deep Learning Applications for Cyber
  Security}}. \bibinfo{publisher}{Springer}, \bibinfo{pages}{45--72}.
\newblock


\bibitem[Xie et~al\mbox{.}(2021)]%
        {congestion2021}
\bibfield{author}{\bibinfo{person}{Xu Xie}, \bibinfo{person}{Chi Zhang},
  \bibinfo{person}{Yixin Zhu}, \bibinfo{person}{Ying~Nian Wu}, {and}
  \bibinfo{person}{Song{-}Chun Zhu}.} \bibinfo{year}{2021}\natexlab{}.
\newblock \showarticletitle{Congestion-aware Multi-agent Trajectory Prediction
  for Collision Avoidance}. In \bibinfo{booktitle}{\emph{{IEEE} International
  Conference on Robotics and Automation, {ICRA} 2021, Xi'an, China, May 30 -
  June 5, 2021}}. \bibinfo{publisher}{{IEEE}}, \bibinfo{pages}{13693--13700}.
\newblock
\urldef\tempurl%
\url{https://doi.org/10.1109/ICRA48506.2021.9560994}
\showDOI{\tempurl}


\bibitem[Xu et~al\mbox{.}(2018)]%
        {xu2018powerful}
\bibfield{author}{\bibinfo{person}{Keyulu Xu}, \bibinfo{person}{Weihua Hu},
  \bibinfo{person}{Jure Leskovec}, {and} \bibinfo{person}{Stefanie Jegelka}.}
  \bibinfo{year}{2018}\natexlab{}.
\newblock \showarticletitle{How powerful are graph neural networks?}
\newblock \bibinfo{journal}{\emph{arXiv preprint arXiv:1810.00826}}
  (\bibinfo{year}{2018}).
\newblock


\bibitem[Ye et~al\mbox{.}(2022)]%
        {GSAN}
\bibfield{author}{\bibinfo{person}{Luyao Ye}, \bibinfo{person}{Zezhong Wang},
  \bibinfo{person}{Xinhong Chen}, \bibinfo{person}{Jianping Wang},
  \bibinfo{person}{Kui Wu}, {and} \bibinfo{person}{Kejie Lu}.}
  \bibinfo{year}{2022}\natexlab{}.
\newblock \showarticletitle{GSAN: Graph Self-Attention Network for Learning
  Spatial–Temporal Interaction Representation in Autonomous Driving}.
\newblock \bibinfo{journal}{\emph{IEEE Internet of Things Journal}}
  \bibinfo{volume}{9}, \bibinfo{number}{12} (\bibinfo{year}{2022}),
  \bibinfo{pages}{9190--9204}.
\newblock
\urldef\tempurl%
\url{https://doi.org/10.1109/JIOT.2021.3093523}
\showDOI{\tempurl}


\bibitem[Yue et~al\mbox{.}(2022)]%
        {yue2022human}
\bibfield{author}{\bibinfo{person}{Jiangbei Yue}, \bibinfo{person}{Dinesh
  Manocha}, {and} \bibinfo{person}{He Wang}.} \bibinfo{year}{2022}\natexlab{}.
\newblock \showarticletitle{Human Trajectory Prediction via Neural Social
  Physics}.
\newblock \bibinfo{journal}{\emph{arXiv preprint arXiv:2207.10435}}
  (\bibinfo{year}{2022}).
\newblock


\bibitem[Zhang et~al\mbox{.}(2022)]%
        {aitp2022}
\bibfield{author}{\bibinfo{person}{Kunpeng Zhang}, \bibinfo{person}{Liang
  Zhao}, \bibinfo{person}{Chengxiang Dong}, \bibinfo{person}{Lan Wu}, {and}
  \bibinfo{person}{Liang Zheng}.} \bibinfo{year}{2022}\natexlab{}.
\newblock \showarticletitle{AI-TP: Attention-based Interaction-aware Trajectory
  Prediction for Autonomous Driving}.
\newblock \bibinfo{journal}{\emph{IEEE Transactions on Intelligent Vehicles}}
  (\bibinfo{year}{2022}).
\newblock


\bibitem[Zhao et~al\mbox{.}(2020)]%
        {zhao2020spatial}
\bibfield{author}{\bibinfo{person}{Xiaodong Zhao}, \bibinfo{person}{Yaran
  Chen}, \bibinfo{person}{Jin Guo}, {and} \bibinfo{person}{Dongbin Zhao}.}
  \bibinfo{year}{2020}\natexlab{}.
\newblock \showarticletitle{A spatial-temporal attention model for human
  trajectory prediction.}
\newblock \bibinfo{journal}{\emph{IEEE CAA J. Autom. Sinica}}
  \bibinfo{volume}{7}, \bibinfo{number}{4} (\bibinfo{year}{2020}),
  \bibinfo{pages}{965--974}.
\newblock


\bibitem[Zhou et~al\mbox{.}(2021b)]%
        {zhou2021sliding}
\bibfield{author}{\bibinfo{person}{Hao Zhou}, \bibinfo{person}{Dongchun Ren},
  \bibinfo{person}{Xu Yang}, \bibinfo{person}{Mingyu Fan}, {and}
  \bibinfo{person}{Hai Huang}.} \bibinfo{year}{2021}\natexlab{b}.
\newblock \showarticletitle{Sliding Sequential CVAE with Time Variant
  Socially-aware Rethinking for Trajectory Prediction}.
\newblock \bibinfo{journal}{\emph{arXiv preprint arXiv:2110.15016}}
  (\bibinfo{year}{2021}).
\newblock


\bibitem[Zhou et~al\mbox{.}(2021a)]%
        {Zhou2021wide_atten}
\bibfield{author}{\bibinfo{person}{Junhao Zhou}, \bibinfo{person}{Hong-Ning
  Dai}, \bibinfo{person}{Hao Wang}, {and} \bibinfo{person}{Tian Wang}.}
  \bibinfo{year}{2021}\natexlab{a}.
\newblock \showarticletitle{Wide-Attention and Deep-Composite Model for Traffic
  Flow Prediction in Transportation Cyber–Physical Systems}.
\newblock \bibinfo{journal}{\emph{IEEE Transactions on Industrial Informatics}}
  \bibinfo{volume}{17}, \bibinfo{number}{5} (\bibinfo{year}{2021}),
  \bibinfo{pages}{3431--3440}.
\newblock
\urldef\tempurl%
\url{https://doi.org/10.1109/TII.2020.3003133}
\showDOI{\tempurl}


\end{thebibliography}

\end{document}